\documentclass[twocolumn]{article}
\usepackage[utf8]{inputenc}

\usepackage[style=authoryear-ibid, dashed=false]{biblatex}
\usepackage{booktabs}
\usepackage{pifont}
\usepackage{multirow}
\usepackage{soul}
\usepackage{color}
\usepackage{hyperref}
\usepackage{graphicx}

\makeatletter
\newcommand{\setword}[2]{
  \phantomsection
  #1\def\@currentlabel{`\unexpanded{#1}'}\label{#2}%
}
\makeatother

\title{Automatic Image Content Extraction:\\ Operationalizing Machine Learning in Humanistic \\Photographic Studies of Large Visual Archives
}

\author{Anssi Männistö$^1$, Mert Seker$^2$, Alexandros Iosifidis$^3$, Jenni Raitoharju$^{2,4}$}

\date{%
    $^1$Unit of Communication Sciences, Tampere University, Finland\\
    $^2$Unit of Computing Sciences, Tampere University, Finland\\
    $^3$DIGIT, Department of Electrical and Computer Engineering, Aarhus University, Denmark\\
    $^4$Programme for Environmental Information, Finnish Environment Institute, Finland\\
}

\begin{document}

\maketitle

\begin{abstract}
Applying machine learning tools to digitized image archives has a potential to revolutionize quantitative research of visual studies in humanities and social sciences. The ability to process a hundredfold greater number of photos than has been traditionally possible and to analyze them with an extensive set of variables will contribute to deeper insight into the material. Overall, these changes will help to shift the workflow from simple manual tasks to more demanding stages. 
    
In this paper, we introduce Automatic Image Content Extraction (AICE) framework for machine learning-based search and analysis of large image archives. We developed the framework in a multidisciplinary research project as framework for future photographic studies by reformulating and expanding the traditional visual content analysis methodologies to be compatible with the current and emerging state-of-the-art machine learning tools and to cover the novel machine learning opportunities for automatic content analysis. The proposed framework can be applied in several domains in humanities and social sciences, and it can be adjusted and scaled into various research settings. We also provide information on the current state of different machine learning techniques and show that there are already various publicly available methods that are suitable to a wide-scale of visual content analysis tasks.
\end{abstract}

\section{Introduction}
\label{sec:intro}

The so called \emph{visual turn} has been one of the major paradigmatic changes in the fields of humanities and social sciences in the past few decades. Usually \emph{visual turn} refers to a shift in emphasis in these fields towards an increasing importance of the visible. It gained prominence in the 1990s and succeeded the \emph{linguistic turn} (\cite[267-278]{jay2002cultural}; \cite[7]{pauwels2000taking}). At the beginning of the 21st century, the visual turn in humanities can be seen as an aspect of another major change, called \emph{digital turn} (\cite{nicholson2013digital}; \cite[194]{wevers2020visual}). These two \emph{turns} - visual and digital - go on at the same time when media companies, museums, archives, and other traditional holders of vast image reservoirs have increasingly brought their collections into public. This opening has provided completely new possibilities for exploring and analyzing  the historical images and visual heritage.

One of the most important prerequisites for building a more visual social science is, as \textcite[211]{grady2007advertising} defines, demonstrating that visual data provide answers to research questions, which are not addressed satisfactorily by the use of more conventional, non-visual, methods. According to \textcite[237]{cerku2019applied}, researchers should be aware of the importance of historical photographs when examining culture, because of their potential to reveal social, economic, and political life of the past. \textcite[2]{emmison2000researching} argue that photographs have been misunderstood as constituting forms of data in their own right when they ``\textit{should be considered in the first instance as means of preserving, storing, or representing information.}" In this sense, photographs should, in their view, be seen as analogous to code-sheets, tape recordings, or any one of the numerous ways in which the social researchers seek to capture data. \textcite[6]{collier2004approaches} states that strong visual research collections present rich potential because they may be productively analyzed in many different ways, both direct and indirect. The ideal situation would be a carefully made visual research collection with comprehensive temporal, spatial and other contextual recording, good annotations, and collection of associated information in an organized data file. However, as \textcite{collier2004approaches} points out, such systematic collections are relatively rare, which leads many scholars to underestimate the true potential of photography, film, and video as reliable sources of cultural information.

Efficient use of visual data in humanities has been hindered by the lack of suitable methodologies for analyzing and accessing large digital image collections. Traditional research methods are based on manual and mostly text-based tools developed in the era of analogue photographs. This limits the amount of considered photographs usually somewhere between 400-1000 (see examples in Section \ref{sec:enhancing}). These numbers are difficult to scale up as each photo must be analyzed individually and manually by a human coder. It is time-consuming to search typical or certain kinds of photos, special patterns or features in photos from a large image collection as the images usually lack content-aware annotations, or this information is incomplete or inadequate.

\textcite[1]{arnold2020enriching} point out that while visual collections may include extensive metadata about the provenance of a digital image, there is often little to no structured data pertaining to the content of the image itself. Similarly, \textcite[195]{wevers2020visual} see that textual analysis in all shapes and forms, has come to dominate digital humanities and, as text-based querying provides mainly textual view of digital sources, ``\emph{as a field, digital humanities has grossly neglected visual content, causing a lopsided representation of all sorts of digitized archives.}" Another problem is that different fields in social sciences, humanities, and behavioral sciences have been developing and using their methods distinct from each other. As \textcite[2]{pauwels2020integrated} states, ``\emph{while visual practices and approaches are being proposed and advocated from myriad disciplinary and theoretical positions, it seems that often more effort is expended in trying to ‘appropriate’ a domain, method, or technique than in developing a more cumulative and integrative stance.}'' 

Many scholars are nevertheless enthusiastic about the possibilities of using machine learning\footnote{Note that we mainly use the term \emph{machine learning}, while \emph{computer vision} could have been an appropriate term as it covers also non-learning-based automatic methods to analyze visual content. However, the main advances discussed in this paper have been made in learning-based techniques, especially in \emph{deep learning}. Thus, also deep learning could be an appropriate term to be used in many places, but our analysis covers a wider variety of machine learning techniques, making machine learning the term of choice.} methods in various fields of social sciences and humanities (e.g., \cite{bodker2018journalism}; \cite{broersma2018exploring}; \cite{fiorucci2020machine}; \cite{sherren2017digital}). \textcite[18]{parry2020quantitative} ponders in her book chapter named \textit{Quantitative Content Analysis of the Visual} in \textit{The Handbook of Visual Research Methods} as follows: ``\emph{Traditional media research techniques have been designed with limited media outlets in mind, manageable via human coding alone. But the ‘big data’ generated by online media practices requires appropriate digital tools and methods, including automated coding or sorting.}" In a same manner, \textcite[9]{pauwels2020integrated}, notes that while ``\emph{the current hype around `big data’ contributes to the idea that empirical research is becoming `more visual', (...) it does not take advantage of the many visual dimensions of culture and society as a prime source of data.}" 

Currently, there are only few examples of the actual use of machine learning tools and methods in these fields and many researchers see machine learning more as a far-future direction than a realistic option for their studies. Rose, who has written one of the most comprehensive books on visual methodologies refers to image content recognition software only briefly in one paragraph (\cite[292]{rose2016visual}). She mentions that researchers are working on programs that will be able to undertake searches based on images' visual content rather than on the tags that have been attached to them. As Rose found such software commercial and very expensive, she chose not to discuss them further in her book. She (\cite[86-87]{rose2016visual}) mentions that practical challenges of accessing and storing images at a large scale are significant. This is why she assumes that most readers of her book will be able to work with a few hundreds of images at most. Surprisingly, machine learning themes are not covered even in a recent comprehensive volume concentrating entirely on methodology, \textit{The Handbook of Visual Research Methods} (\cite{pauwels2020sage}).

Under these conditions, although more and more historical image collections are made public, their utilization does not advance equally fast, unless new methodological tools are put into operation. An obvious reason explaining the rare use of machine learning tools in humanities is the extremely rapid development of the required machine learning techniques during the last decade. In particular, \emph{deep learning} with convolutional neural networks (CNNs) along with the improved computing capabilities have relatively recently opened completely new opportunities in image content analysis (\cite{lecun2015deep}). With the CNNs it is possible to explore the content (what is represented) and the style (how is represented) of images in large archives (\cite[195]{wevers2020visual}) as well as to do \textit{distant viewing}\footnote{\textcite[i6]{arnold2019distant} describe \textit{distant viewing} of large image collections as ``\textit{framework for the automatic extraction of semantic elements of visual materials followed by the aggregation and visualization of these elements via techniques from exploratory data analysis.}" They call for establishing code systems for specific computational interpretations through metadata extraction algorithms. (\cite[i3,i4,i13]{arnold2019distant})} of large collections of visual material before actually studying them (\cite[i3]{arnold2019distant}).

Until recently, as \textcite{wevers2020visual} point out, the computational analysis of visual material has been limited to practical considerations of simple operations, such as adding metadata (\cite{kuhn2018metadata}) or looking at some basic features, such as size, color, or saturation of the visual material (\cite{manovich2012compare}). \textcite[195,203-204]{wevers2020visual} call for new techniques for exploring visual material in archives using non-textual search methods which enable accessing the contents of images directly without referring to textual content, such as titles and captions, and without having to browse through archives manually. They (\cite[198]{wevers2020visual}) state that new machine learning-based techniques contribute also to the construction of a radically new overview of visual culture, which allows for the analysis of trends and changes over extended periods.

During the last few years, pilot studies applying state-of-the-art algorithms for photographic studies have started emerging. \textcite{wevers2020visual} demonstrated the use of a pretrained CNN model for three different automatic visual content analysis tasks over a collection of 450,000 historical images: 1) determining whether an image is an illustration or photograph and defining the historical watershed moment when the number of photographs overtook the number of illustrations, 2) querying images based on abstract visual aspects (clustering visually similar advertisements), and 3) retraining the last layer of the network to classify images into nine relevant visual categories defined by domain experts. A recent study by \textcite{chumachenko2020ww2}, co-authored by most of the authors of this papers, demonstrated the use of CNNs for conducting a novel kind of analysis of prominent Finnish World War II photographers. A CNN was trained to recognize the photographer from the photo, which allowed learning some abstract similarities and differences between the photographers. Furthermore, object detection was applied to analyze most typical topics for each photographer. \textcite{arnold2020enriching} used CNN-based semantic image segmentation to produce structured data on the content of historical photos. \textcite{vaisanen2021exploring} applied state-of-the-art semantic clustering, scene classification, and object detection methods to analyze human-nature interactions in Finnish national parks using visitors' geotagged social media photos as source data. 

\textcite{araujo2020automated} created a protocol for using commercial Application Programming Interfaces (APIs) for Automatic Visual Content Analysis (AVCA) in large-scale image datasets. Traditional visual content analysis (see Section \ref{sec:basics}) starts from particular research questions and hypothesis, which are approached by examining structural content of individual images in a sample. The protocol introduced by \textcite{araujo2020automated}  does not focus on these structural features and related variables and values. Rather, it creates a more qualitative level of analysis: ``\textit{Using our proposed framework, researchers can evaluate the extent to which computer vision labels may be able to assist in the classification of concepts that are theoretically relevant to Communication Science, yet in many cases, the level of complexity of these concepts may fall short.}” (\cite[259]{araujo2020automated}) As \textcite{araujo2020automated} focused on existing APIs instead of the underlying computer vision and machine learning methods, they point out that ``\textit{the black-box approach of commercial API’s also means that these models may change over time, which may pose challenges to replicability}". (\cite[258]{araujo2020automated})

In this paper, we take a more holistic view on the use of machine learning methods for visual studies in humanities and social sciences. We aim to provide a comprehensive picture of how adopting new methods and techniques will enhance traditional settings of photographic studies in various academic fields and of what new approaches these tools enable. We summarize our views into a novel Automatic Image Content Extraction (AICE) framework for large image archives that can replace the previous models when automatic tools are used. In AICE, we reformulate and expand the traditional visual content analysis framework to be compatible with the current and emerging state-of-the-art machine learning tools and to cover the novel machine learning opportunities for automatic content analysis. We also provide information on the current state of different machine learning techniques and show that there are already various publicly available methods that are suitable to a wide-scale of visual content analysis tasks. Our goal with AICE is to provide a general framework for analyzing and searching large image collections. It can be applied in several domains in humanities and social sciences, and it can be adjusted and scaled into various research settings. 

The rest of this paper is structured as follows. Section~\ref{sec:basics} focuses on the basics of traditional visual content analysis. Section~\ref{sec:AICE} introduces our novel Automatic Image Content Extraction (AICE) framework. The following sections illustrate the use of AICE framework by explaining how machine learning can enhance traditional visual content analysis tasks (Section~\ref{sec:enhancing}) and what kind of novel opportunities it will open (Section~\ref{sec:newopportunities}). Section~\ref{sec:risks} discusses the challenges and drawbacks associated with the novel tools. Finally, Section~\ref{sec:conclusion} concludes the paper.

\section{Basics of Visual Content Analysis}
\label{sec:basics}

Content analysis is a method of analysing visual images that was originally developed to interpret written and spoken texts. It was first used in the interwar period by social scientists wanting to analyse journalism of the emerging mass media. It was given boost during the Second World War in order to detect implicit messages in German radio broadcasts. Hence, it is an explicitly quantitative methodology and it was developed to address the need of tackling the sheer scale of the mass media. (\cite[85-86]{rose2016visual}, see also \cite[21-25]{berelson1952content}) 

According to the classic definition of \textcite[18]{berelson1952content}: ``\textit{Content analysis is a research technique for the objective, systematic, and quantitative description of the manifest content of communication.}" While content analysis initially focused mainly on texts, it was also considered in the context of non-verbal information. Nevertheless, it is notable that \textcite{berelson1952content} in his classic book \textit{Content analysis in Communication research} does not even include photographs in his list of ``interesting applications" of quantitative content analysis. Instead, he lists with examples following non-verbal forms of communication: paintings, sculptures, drawings, cartoons, maps, gestures, tone and voice patterns, and music. (\cite[108-109]{berelson1952content})

\textit{Reading Images: The Grammar of Visual Design}, a book by \textcite{kress1996reading}, can be considered as turning point enhancing the use of quantitative methods in studying the contents of photographs in wide variety of fields in humanities and social sciences. In their book, \textcite[1]{kress1996reading} provided ``\textit{inventories of major compositional structures which have become established as conventions in the course of the history of visual semiotics, and to analyze how they are used to produce meaning by contemporary image-makers.}" They called their approach \emph{the grammar of visual design}, which is a large, theory-like concept covering all sorts of images, such as photographs, movies, drawings, paintings, and advertisements. Many scholars have adjusted the theory and its tools to more narrow research settings. \textcite{bell2004content} was maybe the first to use the term \emph{Visual Content Analysis} (VCA) leaning mostly on tools, variables, and values created by Kress and van Leeuwen. In his analysis, Bell studied the differences of cover images of Cleo Magazine published 25 years apart. Bell's version of VCA is a readily operable adaptation of the original grammar and its particularly suitable for photographic analysis. Because of its practicality we have chosen to use it as one of the major sources for the AICE framework we introduce in this paper.

\textcite[8]{bell2004content} describes VCA as ``\textit{a systematic, observational method used for testing hypotheses about the ways in which the media represent people, events, situations, and so on. It allows quantification of samples of observable content classified into distinct categories.}" In a same manner, \textcite[3]{riffe2019analyzing} define quantitative content analysis ``\textit{as the systematic assignment of communication content to categories according to rules, and analysis of relationships involving those categories using statistical methods.}" Although Bell and Riffe speak in the context of media or communication content, there is nothing that restricts the use of VCA only to those research areas.

Lutz and Collins (in \cite[87]{rose2016visual}) separate three factors which favor the use of content analysis in dealing with relatively large numbers of images. First, they suggest that content analysis can reveal empirical results that might otherwise be overwhelmed by the sheer bulk of material under analysis. Second, they insist that content analysis and qualitative methods are not mutually exclusive and content analysis can include qualitative interpretation. Finally, they also suggest that content analysis prevents a certain sort of `bias', because it is a method for analysing images that does not rely on pre-existing interpretative categories.

The factor separating quantitative analysis, like VCA, from many other main research techniques, such as qualitative analysis or discourse analysis, is its focus on the `manifest content'. \textcite[22]{riffe2019analyzing} state that important in this specification is focusing on content's manifest (or denotative or shared) meaning as opposed to connotative (or latent) `between-the-lines' meaning". They stress that quantitative content analysis is the ``\textit{systematic and replicable examination of symbols, which have been assigned numeric values according to valid measurement rules}." The analysis, then, involves comparing relationships of those values using statistical methods. According to \textcite[7-8]{bell2004content}, typical research questions which may be addressed using content analysis include questions of priority/salience or of `bias' of the content and historical changes in modes of representation of, for example, gender, occupational, class, or ethically codified images. As stated by \textcite[58]{emmison2000researching}, one of the principal advantages of quantitative visual research is the historical depth, which can be obtained in an inquiry. 

One of the main challenge of content analysis has been that, if a study focuses on human behavior or visibility in photographs (as the typical cases are), even selecting the photos depicting people is laborious. The process is most burdensome, understandably, with photo archives that are not digitized. With such archives collecting and selecting material and annotating it manually is a very time-consuming practice as researchers must examine each photo individually. This is why it is usually not possible to consider a whole large-scale dataset with the traditional manual methods. With digitized archives, the process is mitigated as researchers can use dedicated software to browse and store images and to use search tools. As search tools usually are text-based, it typically means that only the original caption is used to describe or tell about the content. In such cases, it is not possible to make true content-based searches of the collection. Otherwise, the workflow mode with the digitized archives is quite similar to the traditional process where a researcher actually picks printed photos from a drawer or filing cabinet or uses a microfilm reader. With these traditional workflow modes, a researcher observes features or elements of a photo that are relevant for particular research questions. In quantitative analysis, a researcher or coder writes down certain aspects of a photo to a spreadsheet or database. Typically, these aspects are the amount or qualities of humans or of various object classes. As this observation process is time consuming, these object classes form usually just a fraction of the total amount of features in the photos. 

A fundamental difficulty with visual media is, as \textcite{manovich2013visualizing} see it, that they do not have a standard vocabulary or grammar. This makes the adaptation of automatic visual analysis challenging. The solution \textcite{manovich2013visualizing} suggest is to focus on ``visual form", such as size, color, or saturation, which is easy for computers to analyze, and not semantics, which are harder to automate. However, we believe that systematic semantic analysis of the visual content has potential to provide much more interesting research outcomes and we see the current state-of-the-art machine learning techniques adequate to start using them also for practical semantic analysis purposes. In the next section, we describe our novel AICE framework designed for systematic large-scale visual content analysis using automatic computer vision techniques.

\section{Automatic Image Content Extraction (AICE) framework}
\label{sec:AICE}

Our main purpose is to transform and expand the traditional visual content analysis framework in a way that is compatible with the current and future automatic image analysis methods. We formulate our framework in a similar manner as the version of VCA by Bell (2004) discussed in the previous section, but we significantly enrich it with new variables, values, and connections to existing machine learning techniques. The guiding light for Bell, which we also follow in our framework, was that in order to choose variables, visual content ``\textit{must be explicitly and unambiguously defined and employed consistently (‘reliably’) to yield meaningful evidence relevant to an hypothesis.}" (\cite[9]{bell2004content}). To do that it is ``\textit{first necessary to define relevant \textit{variables} of representation and/or salience. Then, on each variable, values can be distinguished to yield the categories of content which are to be observed and quantified.}" (\cite{bell2004content})  

Our AICE framework is summarized in Tables \ref{tab:avca}-\ref{tab:avca3}. The first two columns form the core of framework. They represent the variables and corresponding values for content analysis considering both the needs of photographic studies and feasibility of automatic analysis. The variables are divided into six main topics: Technical variables (\ref{var:technical}) share some common, basic features of every image and they are useful in many subsequent higher level image analysis tasks (e.g., social distance estimation (\cite{seker2021distance})). With modern photographs, variables 1.2-1.4 can usually be obtained from the metadata, but for historical photos this is typically not the case. When the information is not included, it may also be predicted from the photo content and, therefore, is included in our AICE framework. Especially for the historical photos, the camera technical variables may be also a study question as such to analyze the development of photography. Predicting the photographer from a general photo is naturally almost impossible, but may be possible to some extent when the possible pool of photographers is limited (\cite{chumachenko2020ww2}). The variables in \ref{var:composition} and \ref{var:modality} correspond to the ``visual form'' that is mostly easy to evaluate from the images. The semantic image content and participants, i.e., topics, people, objects, settings, and their properties are included in \ref{var:content}, whereas the interaction between the participants and between the participants and the camera are evaluated in \ref{var:maininteraction}. The variables in \ref{var:similarity}, on the other hand, bring into our framework many important applications of visual analysis based on appearance-based similarity, which may be evaluated focusing on different variables defined in AICE. The practical use of these variables will be illustrated in different examples below. 

\setlength{\tabcolsep}{5pt}
\begin{table*}
    \small
	\caption{Automatic Image Content Extraction (AICE) framework, part 1/3}
	\label{tab:avca}
	\begin{center}
		\begin{tabular}{l|ccc}
	    \toprule
			\textbf{Variable}  & \textbf{Values} & \textbf{Difficulty$^a$} & \textbf{Tasks}\\
			\midrule
			\textbf{\underline{\setword{1 Technical}{var:technical}}}\\
			\setword{1.1 Image type}{var:imagetype}  & Photograph/illustration/map/... & Easy & Classification \\
			\setword{1.2 Camera}{var:camera}$^b$ & Camera model & Difficult & -  \\
			\multirow{2}{*}{\setword{1.3 Focal length}{var:focal}$^b$} & $<$17mm/17-34mm/34-70mm/  & \multirow{2}{*}{Easy} & \multirow{2}{*}{-}  \\
			& 71-300mm/$>$300mm$^c$ && \\
			\setword{1.4 Sensor/film size}{var:sensor}$^b$  & Numerical  & Medium & - \\
			\setword{1.5 Photographer}{var:photographer}$^b$  & Photographer name  & Difficult & Photographer recognition\\
            \midrule
			\textbf{\underline{\setword{2 Composition}{var:composition}}}\\
			\setword{2.1 Photo shape}{var:shape} & Horizontal/vertical/square & Trivial & - \\ 
			\setword{2.2 Photo framing}{var:framing} & Close-up/medium shot/overview & Easy & Framing classification  \\ 
			\setword{2.3 Camera angle}{var:angle}  & Numerical & Medium & Camera pose estimation \\
			
			\midrule
			\textbf{\underline{\setword{3 Modality}{var:modality}}}\\
			\setword{3.3 Color type}{var:colortype}  & Grayscale/color  & Trivial & - \\ 
			\setword{3.4 Tonality}{var:tonality}  & Dark/medium/light & Easy & -  \\ 
			\setword{3.5 Dominant colors}{var:domcolor} & Blue/red/green/yellow... & Easy & -  \\ 
			\setword{3.6 Color distribution}{var:histogram} & Histogram & Easy & - \\ 
			\midrule
			\textbf{\underline{\setword{4 Content/Participants}{var:content}}}\\
			\textbf{\underline{4.1 General}}\\
			\multirow{2}{*}{\setword{4.1.1 Main topic}{var:maintopic}} &  Person/animal/object/   & \multirow{2}{*}{Medium}  & \multirow{2}{*}{Classification}  \\ 
			& environment/event/... & &  \\
			\setword{4.1.2 Salience}{var:saliency} & Salience map  & Easy & Salience estimation \\
			\textbf{\underline{4.2 Persons/actors}}\\
			\textbf{4.2.1 Characteristics} & \\
			\textbf{of single persons} & \\
			\multirow{2}{*}{\setword{4.2.1.1 Status}{var:personstatus}} &  Main character (MC)/   & \multirow{2}{*}{Easy}  & \multirow{2}{*}{Main character recognition}  \\ 
			& side character (SC) & &  \\ 
			
			\multirow{2}{*}{\setword{4.2.1.2 Age}{var:age}} &  Baby/child/young/adult/old  & Easy  & \multirow{2}{*}{Age estimation}  \\ 
			& Exact age & Medium &   \\ 
			\setword{4.2.1.3 Gender}{var:gender} & Male/female/other & Easy & Gender estimation \\
			\setword{4.2.1.4 Identity}{var:identity} & Name & Easy & Face recognition \\
			\setword{4.2.1.5 Ethnicity}{var:race} & European/Asian/African/... & Medium &  \\ 
			\multirow{2}{*}{\setword{4.2.1.6 Height}{var:height}} &  Short/average/tall  & Medium  & \multirow{2}{*}{Height estimation}  \\ 
			& Exact height & Difficult &   \\ 
			\multirow{2}{*}{\setword{4.2.1.7 Weight}{var:weight}} &  Thin/average/fat  & Medium  & \multirow{2}{*}{Weight estimation}  \\ 
			& Exact weight & Difficult &  \\
        \bottomrule			
	\multicolumn{4}{l}{\footnotesize{a) Evaluated by machine learning experts without experimental testing.}}\\
	\multicolumn{4}{l}{\footnotesize{b) This information may be available as metadata, but it can be also estimated from the visual content.}}\\
    \multicolumn{4}{l}{\footnotesize{c) This typology corresponds to superwide/wide/normal angle / medium telephoto / supertelephoto categories}}\\  \multicolumn{4}{l}{\footnotesize{with  cameras that have 'full size' (35mm) digital sensor or use 35mm film.}}\\

		\end{tabular}
	\end{center}
\end{table*}

\setlength{\tabcolsep}{3pt}
\begin{table*}
\small
	\caption{Automatic Image Content Extraction (AICE) framework, part 2/3}
	\label{tab:avca2}
	\begin{center}
		\begin{tabular}{l|ccc}
	    \toprule
			\textbf{Variable}  & \textbf{Values} & \textbf{Difficulty} & \textbf{Tasks}\\
			\midrule
						\setword{4.2.1.8 Occupation}{var:occupation} & Doctor/police/cook/pilot/... & Medium &  \\ 
			4.2.1.9 Role & Child/mother/friend/neighbor/...  & Difficult &  \\
			\setword{4.2.1.10 Nudity}{var:nudity} & Nude/partially nude/clothed & Medium & Nudity detection \\
			\setword{4.2.1.11 Condition}{var:condition} & Healthy/sick/wounded/dead/... & Medium &  \\ 
			\setword{4.2.1.12 Clothes}{var:clothes} & Skirt/trousers/jacket/coat/... & Medium & Clothes recognition \\ 
			\setword{4.2.1.13 Clothing style}{var:clothstyle} & Representative style & Medium & Style clustering\\
			\textbf{4.2.2 Characteristics} & \\
			\textbf{of multiple persons} & \\
			\multirow{3}{*}{\setword{4.2.2.1 Number of people}{var:personnumber}} &  0/1(single)/2(couple)/     & \multirow{3}{*}{Easy}  & \multirow{3}{*}{Person detection}  \\ 
			& 3-6(small group)/7-12(medium group)/ & &  \\ 
			& 13-30(large group)/31-(crowd)$^d$ & &  \\
			\setword{4.2.2.2 Number of groups}{var:groupnumber} & 1/2/3+ & Medium & \\ 
			\multirow{2}{*}{\setword{4.2.2.3 Group typology}{var:grouptypology}} &  Unfocused/common focused/  & \multirow{2}{*}{Medium}  & Gaze estimation,  \\ 
			& jointly focused$^d$ & & social distance est.$^e$ \\
			\setword{4.2.2.4 Group type}{var:grouptype} & Family/friends/sport team/... & Medium & \\ 
			\setword{4.2.2.5 Atmosphere}{var:atmosphere} & Casual/formal/intimate/festive/... & Difficult & \\ 
			\textbf{\underline{4.3 Objects/goals}}\\
			 \setword{4.3.1 Status}{var:objectstatus} & Main motif (MM)/side motif (SM) & Medium &  \\
			 \setword{4.3.2 Text in image}{var:image} & Textual & Easy & Optical character rec.$^e$ \\ 
			 \textbf{\setword{4.3.3 Categories}{var:categories}} & \\
			 \setword{4.3.3.1 Animals}{var:animals} & Cat/dog/cow/horse/fish/... & Easy & Object detection \\ 
			 \setword{4.3.3.2 Common objects}{var:commonobjects} & Book/ball/car/boat/... & Easy & Object detection \\ 
			 \setword{4.3.3.3 Rare object}{var:rareobject} & & Medium & Object detection \\
			 \textbf{\underline{4.4 Settings/events}}\\
			 \setword{4.4.1 Status}{var:eventstatus} & Main motif (MM)/background & Medium &  \\
			 \setword{4.4.2 Indoor/outdoor}{var:inout} & Indoor/outdoor & Easy & Scene recognition \\
			 \setword{4.4.3 Privacy}{var:privacy} & Private/semi-public/public & Medium & Classification \\
			 \setword{4.4.4 Scene}{var:scene} & Urban/rural/forest/hospital/school/...  & Medium & Scene recognition \\
			 \setword{4.4.5 Event}{var:event} & Sport game/wedding/concert/... & Medium & Event recognition  \\
 			\setword{4.4.6 Location}{var:location}$^b$  & Location  & Medium & -\\
			 \setword{4.4.7 Time of day}{var:timeofday}$^b$ & Morning/day/evening/night & Medium & Classification \\
			 \setword{4.4.8 Time of year}{var:timeofyear}$^b$ & Winter/spring/summer/autumn & Medium & Classification \\
			 \setword{4.4.9 Weather}{var:weather} & Sunny/cloudy/raining/snowing/... & Medium & Weather recognition \\
			 \bottomrule	
	\multicolumn{4}{l}{\footnotesize{b) This information may be available as metadata, but it can be also estimated from the visual content.}}\\
	\multicolumn{4}{l}{\footnotesize{d) Categories adopted from (\cite{cristani2020distance})}}\\
	\multicolumn{4}{l}{\footnotesize{e) est. = estimation, rec. = recognition }}\\
		\end{tabular}
	\end{center}
\end{table*}

\begin{table*}
\small
	\caption{Automatic Image Content Extraction (AICE) framework, part 3/3}
	\label{tab:avca3}
	
	\begin{center}
		\begin{tabular}{l|ccc}
	    \toprule
			\textbf{Variable}  & \textbf{Values} & \textbf{Difficulty} & \textbf{Tasks}\\
			\midrule
			 \textbf{\underline{\setword{5 Interaction}{var:maininteraction}}}\\
			 \textbf{\underline{5.1 Spatiality}}\\
			 \setword{5.1.1 Of MC in image plane}{var:mcplane} & Left/right/up/down/middle & Easy & Main character rec.$^f$ \\
			 \setword{5.1.2 Of MC in depth}{var:mcdepth} & Front/middle ground/background & Easy & Main character rec.$^f$ \\
			 \setword{5.1.3 Of MM in image plane}{var:mmplane} & Left/right/up/down/middle & Medium & Object detection \\
			 \setword{5.1.4 Of MM in depth}{var:mmdepth} & Front/middle ground/background & Medium & Object detection \\
			 \setword{5.1.5 Of SC/SM in image plane}{var:scplane} & Left/right/up/down/middle & Medium & Person/object detection  \\
			 \setword{5.1.6 Of SC/SM in depth}{var:scdepth} & Front/middle ground/background & Medium & Person/object detection \\
			 \setword{5.1.7 Spatial relationships}{var:relation} & `X on Y'/`X behind Y'/...  & Medium & Visual relationship det.\\ 
			 \setword{5.1.8 Spatial distribution}{var:spatialdistribution} & Image segmented in classes & Medium & Scene segmentation \\
			 \setword{5.1.9 Spatial coverage}{var:coverage} & Percentage covered by classes & Medium & Scene segmentation \\
			 \textbf{\underline{\setword{5.2 Distances}{var:distances}}}\\
			 \multirow{4}{*}{\setword{5.2.1 Of MC to camera}{var:mctocamera}} & Intimate(-0.5m)/close personal 
			 & \multirow{4}{*}{Medium}  & \multirow{4}{*}{Social distance est.$^f$}  \\ 
			 & (0.5-0.8m)/far personal(0.8-1.2m)/ & &  \\ 
			 & close social(1.2-2m)/far social(2-3.5m)/ & &  \\
			 & public(3.5m- )$^g$ & &  \\
			 \setword{5.2.2 Of MC to others}{var:mctoothers} & As above & Medium & Social distance est.$^f$ \\
			 \setword{5.2.3 Between groups}{var:betweengroups} & As above & Medium & Social distance est.$^f$ \\
			\textbf{\underline{\setword{5.3 Activity}{var:activity}}}\\
			 \setword{5.3.1 Gaze direction of MC}{var:mcgaze} & Towards/away/right/left & Easy & Gaze estimation  \\
			 \setword{5.3.3 Expressions of MC/SC}{var:expression} & Happy/sad/afraid/surprised/... & Medium & Expression recognition \\
			 \setword{5.3.4 Pose of MC/SC}{var:pose} & Standing/sitting/laying & Easy & Pose estimation \\
			 \setword{5.3.5 Action of MC/MM}{var:action} & Jumping/running/falling/...  & Medium & Action recognition\\
			 \textbf{\underline{\setword{5.4 Interaction}{var:interaction}}}\\
			 \setword{5.4.1 Gaze target of MC}{var:mcgazetarget} & Object/person/other & Medium & Gaze estimation \\
			 \setword{5.4.2 Human interaction}{var:huminteraction} & `X hugs Z'/`X talks to Y'/...   & Medium & Human iact. rec.$^f$\\
			 \setword{5.4.3 Human-object interaction}{var:hoi} & `X sits on Z'/`X reads Y'/...   & Medium & Human-object iact. det.$^f$\\
			 
			 \midrule
			
			 \textbf{\underline{\setword{6 Visual similarity}{var:similarity}$^h$}}\\
			 \setword{6.1 Similar images}{var:query} & Images ranked by similarity & Medium & Content-based image ret.$^f$\\
			 \setword{6.2 Appearance-based grouping}{var:clustering}  & Most similar group & Medium & Clustering\\
			 
			\bottomrule	
	\multicolumn{4}{l}{\footnotesize{f) rec. = recognition, est. = estimation, iact. = interaction, det. = detection, ret. = retrieval  }}\\
	\multicolumn{4}{l}{\footnotesize{g) The original distances by \cite{hall1966hidden} were in feet. We converted to the metric system and rounded to the nearest 0.1m for}}\\ \multicolumn{4}{l}{\footnotesize{closer categories and to the nearest 0.5m for further categories.}}\\
	\multicolumn{4}{l}{\footnotesize{h) Similarity may be based on any of the above defined variables, e.g., visual form, face, or objects.}}\\
		\end{tabular}
	\end{center}

\end{table*}

It should be taken into account that the variables are interconnected. Machine learning methods for more complex tasks can need other variables, e.g., to perform age estimation, the people in the images must be detected first. In some cases, the same machine learning method (e.g., a trained CNN) may evaluate several variables (e.g., height and and weight estimation) or partial combinations of different variables (e.g., a classifier can categorize both scenes and events). The values suggested in the second column are only advisory. After selecting the variables of interest, the target values must be decided based on the research goals (and constrained by available machine learning techniques).

The third column (`Difficulty') aims at giving an impression on the difficulty of the tasks for the current automatic image analysis and machine learning algorithms. `Trivial' stands for variables that can be easily and without any errors determined from the image, `Easy' tasks can be performed using the existing algorithms with a high accuracy, but there may be some errors or ambiguities. `Medium' tasks can be currently performed to some extent, but errors are expected. Finally, `Difficult' tasks may remain partially unsolvable also in the future. For example, professions may be recognized in some special cases where distinctive clothing or equipment is used, while defining the profession of a random person walking on a street during his/her free time is almost impossible. It should be noted that the evaluation is based on subjective views of the co-authors having a background in machine learning research. No experimental comparisons of the difficulty level were conducted. The difficulty levels depend significantly on the source images (e.g., the time period and environment where the photos were taken) and are subject to change as new algorithms for different tasks are being developed. Furthermore, it should be noted that we evaluated the difficulty levels considering only the difficulty of the task itself assuming that large-scale annotated training datasets are available. However, the more specialized the analysis task at hand, the less likely it is to find suitable training dataset or the training sets may contain only a small subset of topics of interest. We return to the challenges of the training data in Section \ref{sec:risks}. 

Finally, the last column (`Tasks') links the visual content analysis variables with the available methods and terminology used in the computer vision field. The description of the specific methods available for different tasks is beyond the scope of this paper, but interested readers may refer to articles in computer vision and machine learning fields to look for more information. A list of state-of-the-art methodologies for the different problems is: classification (\cite{krizhevsky2012imagenet}), photographer recognition and framing classification (\cite{chumachenko2020ww2}), camera pose estimation (\cite{kendall2017camerapose}), basic color analysis (\cite{gonzalez2018digital}), salience estimation (\cite{li2015saliency}), main character recognition (\cite{seker2021maincharacter}), age and gender estimation (\cite{rodriguez2017age}), face recognition (\cite{wang2021face}), height and weight estimation (\cite{altinigne2020height}), nudity detection (\cite{ion2019nudity}), clothes recognition (\cite{liu2016fashion}), clothing style clustering (\cite{matzen2017streetstyle}), person detection (\cite{braun2019eurocity}), gaze estimation (\cite{kellnhofer2019gaze}), social distance estimation (\cite{seker2021distance}), optical character recognition (\cite{memon2020handwritten}),  object detection (\cite{liu2020detection}), scene recognition (\cite{zhou2017places}), event recognition (\cite{wang2018events}), weather recognition (\cite{zhao2018weather}), scene segmentation (\cite{fu2019segmentation}), expression recognition (\cite{li2020expression}), pose estimation (\cite{cao2019openpose}), action recognition (\cite{dong2021action}), visual relationship detection
(\cite{yu2017relationship}), human interaction recognition (\cite{stergiou2019interaction}), human-object interaction detection (\cite{li2019hoi}), content-based image retrieval (\cite{tzelepi2018cbir}), and clustering (\cite{min2018clustering}).

We can now return to the recent pilot works demonstrating the use of CNN models for photographic studies described in Section~\ref{sec:intro}. The first task considered by \textcite{wevers2020visual} was to classify images into illustrations and photographs, which corresponds directly to the variable \ref{var:imagetype} of the AICE framework. In the second task, they clustered visually similar advertisements, which corresponds to the AICE variable \ref{var:clustering}. In the third task, they classified images into nine classes: buildings, cartoons, chess, crowds, logos, maps, schematics, sheet music, and weather reports. This task can be seen as a combination of the variables \ref{var:imagetype}, \ref{var:maintopic}, and \ref{var:categories}. \textcite{chumachenko2020ww2} aimed at evaluating the photographer based on the photo, i.e., the variable \ref{var:photographer}. Furthermore, they applied object detection to locate common objects and animals in the photos according to the variables \ref{var:animals} and \ref{var:commonobjects} and evaluated the variable \ref{var:framing} by categorizing the photographs into close-ups, mediums shots and overview.
\textcite{arnold2020enriching} segmented historical photos into regions containing elements such as sky, water, trees, grass, and roads according to variable \ref{var:spatialdistribution} to produce structured data on the content of historical images. \textcite{vaisanen2021exploring} used scene classification (\ref{var:scene}), object detection (variables \ref{var:animals} and \ref{var:commonobjects}), and semantic clustering (\ref{var:clustering}) to analyze human-nature interaction in Finnish national parks. They analyzed typical photo content and differences between national and international national park visitors.
In the next two sections, we provide more examples of potential uses of our AICE framework by discussing how automatic image analysis could have enhanced some traditional visual content analysis studies and what novel opportunities it can open. 

\section{How machine learning can enhance traditional visual content analyses}
\label{sec:enhancing}

In this section, we have collected some pioneering studies that have used relatively large collections of images - photos the researchers themselves have taken, advertisements and media images, photos stored in various photographic archives - in many fields of humanities and social sciences. We describe the material collection and the research methodology used and elaborate how applying machine learning tools according to the proposed AICE framework could have advanced the process of collecting material or determining variables and their values, and, thus, possibly enhanced the results. List of studies selected here is, of course, by no means comprehensive.

In traditional photographic research settings of media images using visual content analysis, the number of photographs is typically between 400-1000. As examples of war or crisis related topics, \textcite{griffin1995picturing} constructed a systematic inventory of the types, range, and frequency of photographic images presented in American news magazines during Persian Gulf War with 1104 images (also \cite{griffin2004picturing}), \textcite{mannisto2004luottamus} compared the topics of photographs of Afghanistan War (2001, 204 photos) and Iraq War (2003, 343 photos) in Finnish newspapers, \textcite{morse2014covering} studied death images in Israeli newspapers with 506 photos, and \textcite{wilmott2017politics} explored how Syrian refugees have been visually portrayed in the U.K. online media by employing a visual content analysis of 299 photographs. These studies share some basic variables that are core in our AICE framework, such as \ref{var:personnumber}, \ref{var:gender}, \ref{var:age}, \ref{var:framing}, \ref{var:personstatus}, and \ref{var:saliency}. In all these studies, similar to all the other studies discussed later in this section, one of the main benefits of using automatic image analysis would be the possibility to conduct the study on a much larger dataset, possibly collected from additional sources and time periods.     

One of the pioneering projects using photographic information in the field of visual anthropology was `Balinese Character: a Photographic Analysis' by Bateson and Mead (1942). The study was carried out in the late 1930s with the aim of inquiring Balinese culture. In the center of the book, there are 759 photographs organized in different themes. The original amount of photos was 25,000. The analysis focused on gestures and bodily expressions (\cite{emmison2000researching}). Machine learning tools for AICE variables \ref{var:pose}, \ref{var:action}, \ref{var:interaction}, and \ref{var:hoi} could be used to automatically repeat a similar analysis on the full dataset. 

\textcite{goffman1979gender} analyzed stereotyped portrayals of males and females in magazine advertisements in a seminal work `Gender Advertisements' including 508 images. After Goffman's work, several researchers have extended the inquiry into gender stereotypes using replications of Goffman's content analysis methodology, as stated by \textcite[205]{bell2002goffman} (See also \cite{emmison2000researching}). For our paper, it is beneficial to take a closer look on the way \textcite{bell2002goffman} revisit Goffman's work. Their goal was to demonstrate usefulness of combining quantitative analysis with semiotic approach. In order to do this, they developed further Goffman's approach utilizing dimensions of visual structure, derived from and operationalized using Kress and van Leeuwen's system of analysis (see Section~\ref{sec:basics}). They used this apparatus to analyze 827 advertisements collected from popular Australian magazines published during 1997–8. (\cite{bell2002goffman})

Many of the hypotheses \textcite{bell2002goffman} include are relatively easy to analyze automatically. Examples of such hypotheses are: ``\emph{In terms of social distance, females will be framed more intimately and be less likely to be represented at a ‘public’ distance than men}", ``\emph{Women, more frequently than men, will be shown on the right section of a layout that is structured along the horizontal axis}", and ``\emph{Women will less frequently be depicted gazing at the camera/viewer than men}". AICE variables \ref{var:gender}, distance of \ref{var:mctoothers}, \ref{var:mcplane}, and \ref{var:mcgaze} directly correspond to and Bell and Milic's interests, whereas group  formation of genders could be examined by combining methods for \ref{var:groupnumber}, \ref{var:grouptypology}, and \ref{var:grouptype} with gender recognition. One dimension \textcite{bell2002goffman} give a lot of attention is the formal relations between depicted elements which can be analyzed in terms of position in the frame. The AICE framework is particularly capable acknowledging the spatial distribution of persons and objects inside the frame both in image plane and in depth. Several AICE variables, such as the spatiality of the main character (\ref{var:mcplane}) and the main motif (\ref{var:mmplane}) in image plane or in depth (\ref{var:mcdepth} and \ref{var:mmplane}), will contribute to this. With machine learning tools it would we easy to add also other related variables such as \ref{var:personstatus}, \ref{var:saliency}, \ref{var:age}, or \ref{var:action}.

\textcite{kozol2009life} began in 1980s to research Life magazine, which is considered as the most influential publication of visual news in the post Second World War period. She wanted to analyze whether the traditional “family values” were referenced as an idealized portrait of the white, middle-class nuclear family consisting of female housewives and male breadwinners. While collecting the material, Kozol faced several challenges. The first and foremost problem was the amount of material. There were simply too many issues to examine. In the end, Kozol selected to her study every issue in the months of October and May from fifteen years of the post war period looking for news photo-essays that included pictures of families. Kozol’s experience is a typical example where a researcher cannot study the large amount of material as a whole. It is also a good example where machine learning tools could enhance the process. Machine learning could both filter out the potential family photos by selecting photos containing a woman, man and children (AICE variables \ref{var:personnumber}, \ref{var:age}, \ref{var:gender})  and be used to carry out also further analysis including \ref{var:race} and \ref{var:occupation}. While occupation may be difficult to define automatically (also manually), housewife detection could be eased by detecting actions (\ref{var:action}) such as cooking, cleaning, or childcare.

Lutz and Collins (in \cite[87-93]{rose2016visual}), made content analysis of nearly 600 of the photographs published in National Geographic between 1950 and 1986. They collected the sample choosing one photo at random from each of the 594 articles on non-Western people. This procedure produced a manageable number of photos that Lutz and Collins and their research assistants could analyze manually. In a close reading of those photos, they examined issues of race, gender, privilege, progress, and modernity through an analysis of factors such as color, pose, framing, and vantage point used in representations of non-Western peoples. (\cite{1993reading}). Lutz and Collins (in \cite[93]{rose2016visual}) coded each photo with 22 categories (instead of term `category' we would use here the term `variable'). Most of the variables they used are directly included in or are combination of variables in AICE. At least the following variables are included: smiling in a photograph (\ref{var:expression}), gender of adults depicted (\ref{var:gender}, \ref{var:age}),  age of those depicted (\ref{var:age}), activity type and level of main foreground figures (\ref{var:personstatus},\ref{var:pose}, \ref{var:action}), camera gaze of main person (\ref{var:personstatus},\ref{var:mcgaze}), surroundings of people and urban versus rural setting (\ref{var:inout}-\ref{var:scene}), group size (\ref{var:personnumber}), male/female nudity (\ref{var:gender}, \ref{var:nudity}), and vantage point (\ref{var:angle}). If Lutz and Collins had been able to utilize machine learning based automatic method for analysing the content, it would have enabled them to use a much larger sample, e.g., all the several thousands of photos that were in the original 594 articles. While some of the included tasks cannot be currently performed automatically at human accuracy, the larger sample would still give statistically significant results and more variables could be easily added.

\textcite[211]{grady2007advertising} made a close examination of depictions of black persons in advertisements published in Life magazine, 1936-2000. His study is a good example of a very demanding research setting in which a carefully developed quantitative data forms a ground for making interpretations of some sensitive questions, such as how the commitment of white population to racial integration has changed. Grady's sample consisted of nine chronologically stratified periods from 1936-2000 and contained a total of 590 advertisements, which were scanned and entered into a database. The coding was done manually on a spreadsheet. Two among the most central observations in Grady's study (\cite[225,234]{grady2007advertising}) are very interesting as they are combinations of variables in the proposed AICE framework. The first observation was on the percentage of black persons depicted in personal or private settings. This could be operationalized in AICE using variables of \ref{var:race} and \ref{var:privacy}. The other observation was on the percentage of black persons in the sample where there is eye contact between black and white persons. The values in this variable were ``Eye contact between races" and ``One race looking at another". Grady noticed that his analysis shows that in the iconography of segregation black persons tended to look at whites admiringly, while whites looked at blacks in a more patronizing and condescending fashion. Today, Grady concludes, there is not much difference in how these two ethnic groups look each other; it is just that they hardly do so in advertisements. A part of a similar analysis could be conducted by combining AICE variables of \ref{var:race} and \ref{var:mcgaze}. However, evaluating whether a look is admiring or patronizing would be very challenging. Even though the task can be seen as evaluating the variable \ref{var:expression}, it is an example where the selected target values make the task very difficult. Furthermore, annotated large-scale datasets for expression recognition do not contain these categories meaning that a lot of data should be first manually annotated to be able to even try training a machine learning model.   

Of the above-mentioned studies, we take Wilmott's paper for closer examination. It was published in 2017 and was thus made in a time when many machine learning tools would have been already available but were not used. The paper is noticeable and rare as it brings together a detailed VCA setting of a very serious theme and the use of Securitization Theory, which is a special branch of study of international relations. In that context, \textcite[68]{wilmott2017politics} focused on \textit{visual communication acts}, or what she refers to as \textit{picture acts}. Two of the three research questions Wilmott sought to answer were such that they could benefit on the use of AICE: 1) How does gender figure into the visual depictions of Syrian refugees? 2) How do the visual depictions of Syrian refugees differ across U.K. online newspapers? 

The U.K. online quality news websites in Wilmott's study were The Guardian, The Telegraph, and The Independent. Thus, the sample reflects three different affiliations in the political spectrum. Every newspaper article in a certain time frame was scanned, and all the photographs of Syrian refugees were collected and included in the study according to several criteria, finally yielding inclusion of 299 photographs. Several coders annotated the material utilizing a codebook created originally by the Media Department of London School of Economics. (\cite[71]{wilmott2017politics})

The research was built on several different technical considerations to evaluate whether photographs humanize or dehumanize refugees (\cite[72]{wilmott2017politics}). Most of these considerations could be easily automated following the AICE framework. Wilmott's first interest was to examine whether refugees are portrayed as individuals. This was done because previous studies assert that photographs of individual victims evoke more empathetic emotions in viewers than groups of victims, and further that large groups overshadow individuality and attribute common characteristics to all members of a given social group. Following preceding examples, Wilmott coded images of refugees into four categories, from (1) individuals to (4) large groups (AICE variable \ref{var:personnumber}). The content analysis was then further refined through an experiment that explored a second factor: camera distance, using traditional categories of frame of view, starting from various types of `close-ups' and ending to various `overview shots' (\ref{var:framing}). Moreover, coding also looked at whether the refugees look directly into the camera (\ref{var:mcgaze}). These decisions were based on literature that suggests that closer photos, especially those in which subjects look directly into the camera, produce a more intimate association with refugees. (\cite[72]{wilmott2017politics}) Furthermore, as \textcite[72-73]{wilmott2017politics} describes, men are more often associated with violent behavior than women, and a group of female migrants more effectively visualizes the feeling of threat. Therefore, photographs have been coded by `male and female', `female', `male', `only children depicted', and `unclear' (\ref{var:gender} and \ref{var:age}). Wilmott's findings indicate that the image of the refugees has become mainstreamed and that there were no major differences on representing refugees between the compared newspapers. \textcite[77-78]{wilmott2017politics}

As the number of the photos in the study is rather low, only 299, it is obvious that the analysis of such a small sample would not be ideal or even proper use of machine learning tools at all. However, we note that currently the results have not been compared to the average images in the investigated newspapers and, therefore, the conclusion of the image of the refugees becoming mainstreamed lacks some evidence that could be obtained by comparing with the actual mainstream images. Furthermore, while no significant differences between the newspapers were found, comparisons against the average photos in the respective papers could show whether this differs from the typical images and whether there are differences between the newspapers from this point of view. We will get back to this example soon in the following section. 

\section{Potential new opportunities in photographic studies brought by machine learning}
\label{sec:newopportunities}

\subsection{Large-scale multi-factor comparisons}

While we have mentioned already several times that machine learning methods are especially suitable for very large data-sets, it should be emphasized that this especially concerns comparisons. Machine learning methods tend to do some errors. For example, it could be that a distance estimation algorithm tends to somewhat underestimate the distance of the main character to the camera or a person detection algorithm might find only 95\% of the people in the photos. However, these sort of errors can be considered systematic as long as the considered sample is large enough and the photos to be compared have a similar quality. Therefore, claims such as ``Photo collection A contains exactly 4521 persons" or ``In photo collection B, the average distance of the main character to the camera is 2.3m" may not be feasible, but the differences between datasets can be more accurately estimated, e.g., as ``In photo collection A, photos have on average 30\% more people than photos in collection B."

We now return to Wilmott's study (\cite{wilmott2017politics}) on depicting Syrian refugees in newspapers, which we described in the previous section. AICE could bring a totally new layer of analysis as it could annotate \textit{all} the photographs in a certain period in the three newspapers, not just the 299 belonging to articles concerning Syrian refugees. By annotating the whole material of thousands of images, the machine learning enhanced analysis could create a reliable reference material. Wilmott's study did show that the differences between the newspapers were small, but it did not tell how the samples compared in relation to the typical ways each of the newspapers represent the qualities of persons in photographs. Consider, for example, The Independent. What if it turned out that it does in general have a particularly low number of portrays with eye contact, while at the same time that number would be very high in the other two newspapers in question. In this imaginary case, the relative difference of depicting the eye contacts of Syrian refugees compared to the general way of showing eye contact in person photos would be considerably different between the newspapers. This is of course just a hypothetical example, but nevertheless illustrates how the ability to use the entirety – not just a fragment – of photos in the analysis would significantly broaden the analysis.

Another important aspect is that AICE allows to easily include multiple variables into the research setting and look for any interesting correlations between them. It is understandable that using the traditional methods no one wants to invest a large amount time to manually annotate some variables that are unlikely to be interesting for the study at hand. However, with machine learning-based methods such analysis can be carried with little extra effort. While in many cases this may indeed confirm that the variable is not interesting, it is also likely that at some point such trials lead to interesting observations and novel insights of some phenomena that would have never been revealed by the traditional research methodology.

\subsection{Spatiality and use of space}
\label{ssec:spatiality}

In our view, two interrelating dimensions can be separated in spatiality and use of space in photographic studies: (1) positions of persons and objects inside the frame both in the image plane and in depth and (2) social distance of participants in an image. Analyzing spatiality has been an important goal in visual research ever since \textcite{hall1966hidden} published his theory of \emph{proxemics} and social distances. Spatiality is an important aspect also for \textcite{emmison2000researching}, who are the authors of the book \textit{Researching the visual}, considered as one of the most comprehensive introductions to visual inquiry. According to \textcite[4]{emmison2000researching}, objects, people, and events which constitute the raw materials for visual analysis, are not encountered in isolation but rather in specific context: ``\textit{It is this spatial existence which serves as the means whereby much of their sociocultural significance is imparted. Visual data, in short, must be understood as having more than just the two-dimensional component.}" They (\cite[191]{emmison2000researching}) describe that everyday public behaviour has a spatial and territorial component to it. Their thinking leans on some classical texts in sociology, such as Simmel's `The Sociology of senses' from early 20th century. Simmel drew attention to one of the key features of modern life, that interaction is based on sight rather than sound. (\cite[192]{emmison2000researching})

The study of the ways humans utilize or orient to space described by Hall is essential also for the theory of \textcite{kress1996reading}. According to Hall, in everyday interaction, social relations determine the distance (literally and figuratively) we keep from one another. We carry with us a set of invisible boundaries beyond which we allow only certain kinds of people to come. The location of these invisible boundaries is determined by configurations of sensory potentialities – by whether or not a certain distance allows us to smell or touch the other person, for instance, and by how much of the other person we can see with our peripheral (60 degree) vision. (\cite[24]{bell2004content}; \cite[130-131]{kress1996reading}) 

At the core of Hall's theory is a typology with categories corresponding different fields of vision. At \textit{intimate} distance, we see the face or head only. At \textit{close personal} distance we take in the head and the shoulders. At \textit{far personal} distance we see other person from the waist up. At \textit{close social} distance we see the whole figure. At \textit{far social} distance we see the whole figure ‘with space around it’. And at \textit{public} distance we can see the torso of at least four or five people. (\cite[114-129]{hall1966hidden}. See also \cite[24]{bell2004content}; \cite[129-31]{kress1996reading}) For Kress and van Leeuwen, equally important as the social distance is the use of space in terms of positions of persons or objects within the frame. They show, for example, that the value of an element in the top half of a portrait-shaped frame is different from its value if located in the bottom half. A similar difference in information value exists also between the dimensions of center and margin in an image structure.  (\cite[193-208]{kress1996reading}. See also \cite[211]{bell2002goffman}).

While exploring the changes and trends in the ways humans utilize space in everyday interaction is at the core of many historical and sociological studies using photographic archives as source material, proper tools for finding and measuring spatial qualities effectively have been lacking. Acknowledging and operationalizing them with machine learning tools would mean a major leap in photographic studies. Basically, this could enable to consider 2D photos as representations of a certain 3D situation or scene. By doing that, machine learning has potential to revolutionize many traditional research settings, especially those interested in the interaction of people. 

As shown in Table \ref{tab:avca3}, we have adopted the above-described proximity typology as suggested target values for distance-related variables under \ref{var:distances}. This typology can be utilized at least in three ways in photographic studies, all of which could be operationalized with different machine learning tools. First, it will tell also about the physical distance between the photographer and the main character or theme (\ref{var:mctocamera}). This distance unavoidably affects the interaction between the photographer and the main character. Second, determining the mutual distances of persons in photos (\ref{var:mctoothers}) will give information of what kind of relations the persons have with each other, are they close or far away from each other, how are they interacting, and so on. Third, the typology can be used to evaluate group formation (\ref{var:groupnumber}) and distances between groups (\ref{var:betweengroups}). 

Furthermore, the fields of vision linked with the proximity typology closely correspond to the traditional definitions of photo framing in film and television. Framing is also a basic property and feature of every photograph, telling if it is a close-up or an overview (or something in between) of the main character or theme. When this information is automatically determined (AICE \ref{var:framing}) and included in the metadata of a photo, it will serve as one of the basic variables in searching images from large image archives in many professional use cases. Photo framing information can be accompanied by information on camera angle (\ref{var:angle}). To bring out more aspects of the three-dimensional situation that took place in the front of the camera, the variable \ref{var:relation} focuses on how the participant locations are linked to each other (e.g., under/next to/behind/in front of). 

To give a more practical example, we turn our attention to a study of \textcite{temelova2011daily} on daily street life in Prague. Their basic questions were ``Who are the users of the space?" and ``Are there spatial and temporal differences in the manner various social groups use the public space?". In order to demonstrate dimensions of differentiation, they explored the heterogeneity of users based on differences in people's wealth, age, ethnicity, and everyday practices. Methodologically, their study relied primarily on the fieldwork and direct observation of incidents and events which characterise the outdoor life of the area. Photographs were used only to illustrate typical and special ways of utilizing public space and peoples mutual distances and also to identify people belonging to various categories (age, professions, etc.). 
The kind of research setting Temelová and Novák carried out would be possible to create, e.g., by using large amount of photographs taken from a fixed position with certain intervals of time. Along with the above-discussed variables related to spatiality, it would be easy to automatically analyze other variables, such as \ref{var:age}, \ref{var:gender}, \ref{var:pose}, and \ref{var:action}, Temelová and Novák were interested in. Others, such as whether a person belongs to the category of ``managers and professionals", estimated in terms of his/her clothing, of course, are more challenging, but may be approximated using variables \ref{var:occupation} and \ref{var:clothes}. This kind of approximation is present also in the original research setting using direct observation. Moreover, if specific machine learning tools are developed and finetuned to the settings like this, a similar research could be then easily repeated in many cities or places as a comparative approach.

It should be noted that the differences and trends related to proxemics may naturally be quite subtle and not possible to extract from a small set of images. Instead, very large datasets are needed to reliably evaluate average behavioral traits. Such research has not been feasible using traditional manual research methods and, thus, machine learning methods may open a novel unexplored research direction. Analyzing photographic contents of the news media during the COVID-19 crisis, for example, would provide valuable material for observing the changes in peoples behavior in terms of their use of personal space. During this “corona era”, the public speech was filled with notions concerning how far away people should be from each other. Reporting of the crisis filled or affected practically the whole coverage of the news media and the time span was rather long. It would form, thus, a good entity against which to compare findings in different countries or of a reference period few years earlier.

\subsection{Analysing interaction, gestures, and gaze}
\label{ssec:actions}

Analysing interaction, gestures, postures, touching, or gazes between persons in photos is a typical task in many photographic research settings. As \textcite[218-220]{emmison2000researching} put it, humans use body language consciously or unconsciously to convey information to others, e.g., about emotions, social status, openness to interaction, and sexual arousal. Much of the most interesting theory is on the role of gaze and eye contact in contemporary social life. However, as the authors note, such work is difficult to conduct, because of the unobservability of the gaze. As Simmel and others have observed, the modern city has brought people into proximity. This has caused problems for the use of gaze and also, on occasion, for touch (\cite[225]{emmison2000researching}). With traditional visual inquiry methods, analysing these kinds of variables from the photographs is one of the most time-consuming phases of any research process as it needs a lots of interpretation.

Machine learning offers several tools for automating this kind of analysis shown in AICE under variable categories \ref{var:activity} and \ref{var:interaction}. Here, it should be noted that while evaluating a person's pose (standing/sitting/laying) is relatively easy, it is challenging to evaluate actions with movement as a single shot of, e.g., walking, running, or jumping person, may look very similar and more information (video or several shots) is needed for reliable estimation. On the other hand, tools for \ref{var:query} can be used to find out whether an interesting gesture or pose appears repeatedly, and tools for \ref{var:clustering} can be used to divide action and poses automatically into groups containing similar gestures (see Section \ref{ssec:appearancebased}). 

Combining this data (what he/she is doing and to where he/she is looking at) with scene recognition (e.g., \ref{var:scene} and \ref{var:event}) and with the spatial information as discussed in \ref{ssec:spatiality} will produce an unprecedented and rich compilation of content properties of an image. These data are useful as such when making automatic annotations of photographic contents for various needs of image archives. It enables also constructing new kinds of search tools and methods and it will, thus, also contribute to improving accessibility of the image archives. The tools would be useful in several fields of social sciences and humanities, studying, e.g., possible changes of external forms of, or interaction between, populations in certain time span.

\begin{figure}[t]
    \centering
    \includegraphics[width=1\columnwidth]{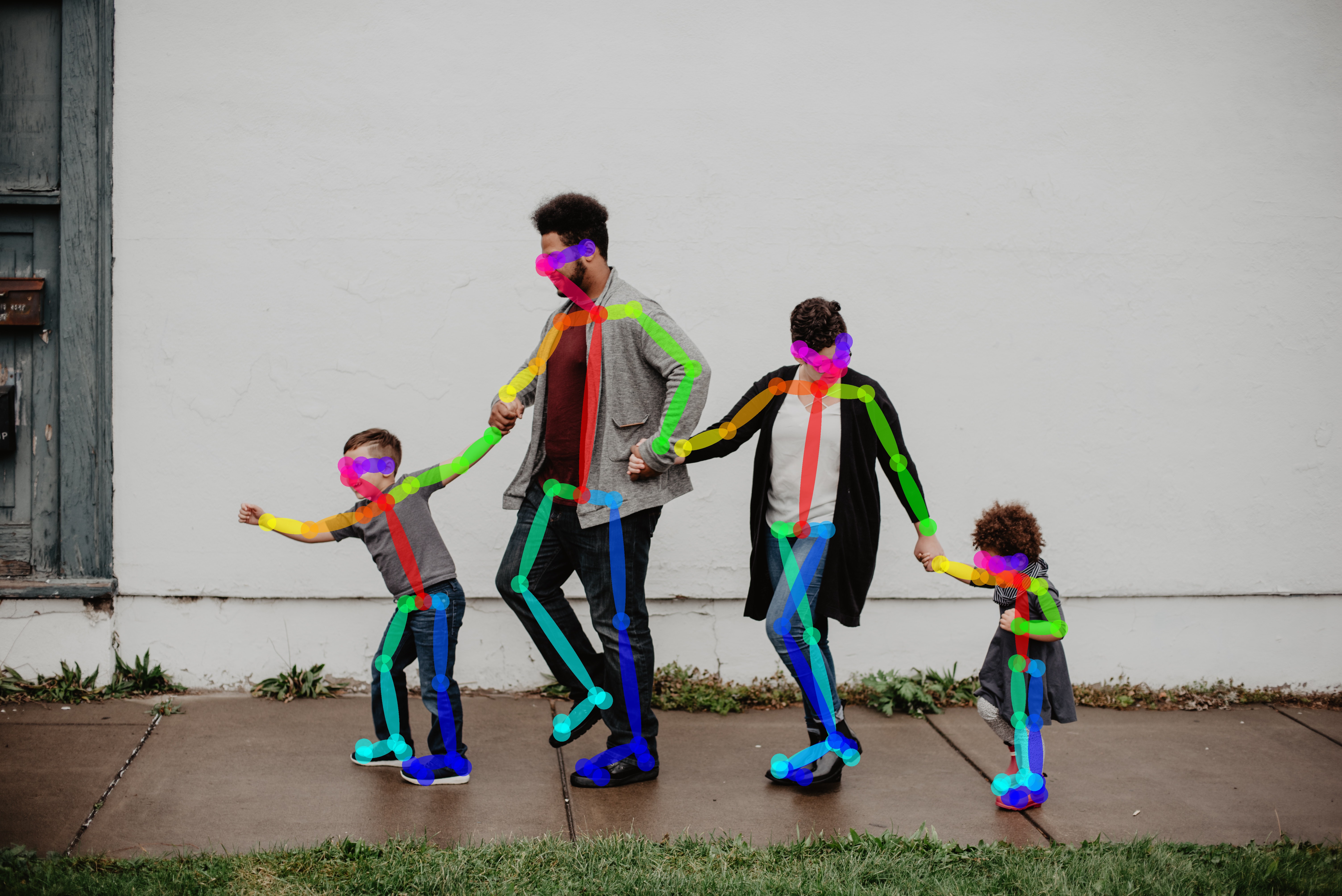}
    \caption{An example of a skeleton model (\cite{cao2019openpose}) that can be used for automatic pose analysis. (Image source: \textit{https://www.pexels.com})}
    \label{fig:openpose}
\end{figure}

As a practical example, pose estimation, such as the OpenPose model (\cite{cao2019openpose}) illustrated in Fig. \ref{fig:openpose}, could be used in a sociobiological study similar to the study Klein carried out searching for differences in sitting postures of males and females. In Klein's study, postures were recorded in 600 men and women and the study focused on typical ways of sitting, depending on gender, age, and place, in one city in 1984. (\cite[220-226]{emmison2000researching}) With machine learning tools it would be possible to classify not just the original material, but reference material from \textit{other} cities and in \textit{other} times, depending of course on the available archives. Furthermore, automatically clustering sitting poses according to the AICE variable \ref{var:clustering} could reveal typical sitting pose variants that were not observed in the manual examination.

\subsection{Salience}

Defining what is the most salient element is present in almost every research setting in humanities and social sciences using photographs as source material. The most salient elements are usually also those which at minimum need to be acknowledged and annotated when describing the content of a photograph. As shown earlier in this paper, human behavior or visibility is typically at the core of study in many research settings. In such cases, the researcher usually needs to define who is the main character, which in most cases is also the most salient participant in the photograph. Defining salient elements can, thus, be considered as one central functions in visual content analysis.

As \textcite[21]{bell2004content} describes, content analysis can show what is given priority or salience and what is not. It can show how images are connected, who is given publicity and how, as well as which agendas are run by particular media.
\textcite[179-220]{kress2021reading} discuss salience in detail with examples of many types of images (movies, classical art, advertisements, and magazine page layout). For them, salience indicates that \textit{ ``some image elements are made more conspicuous than others elements (…) and this makes them more important of information in the whole"} (\cite[181]{kress2021reading}).  Salience is also one of the three key compositional principles of images alongside framing and information value of different parts of an image (\cite[204]{kress2021reading}). 

Traditionally, detecting salience from images has required a lot of attention – and interpretative work from the researcher – as salience can be a result or a combination of several different variables. This becomes obvious when looking how \textcite[211]{kress2021reading} define the term: \textit{``Visual salience results from a complex interaction (…) between a number of factors. These include size, sharpness of focus, (…) areas of high tonal contrasts (…), color contrasts, placement in the visual field, perspective (foreground objects are more salient than background objects, and elements that overlap other elements are more salient than elements they overlap)"}. Also, as  \textcite[216]{kress2021reading} point out, it is not only what is made salient, but also how it is made salient that contributes to the meaning.

\begin{figure}[t]
    \centering
    \includegraphics[width=1\columnwidth]{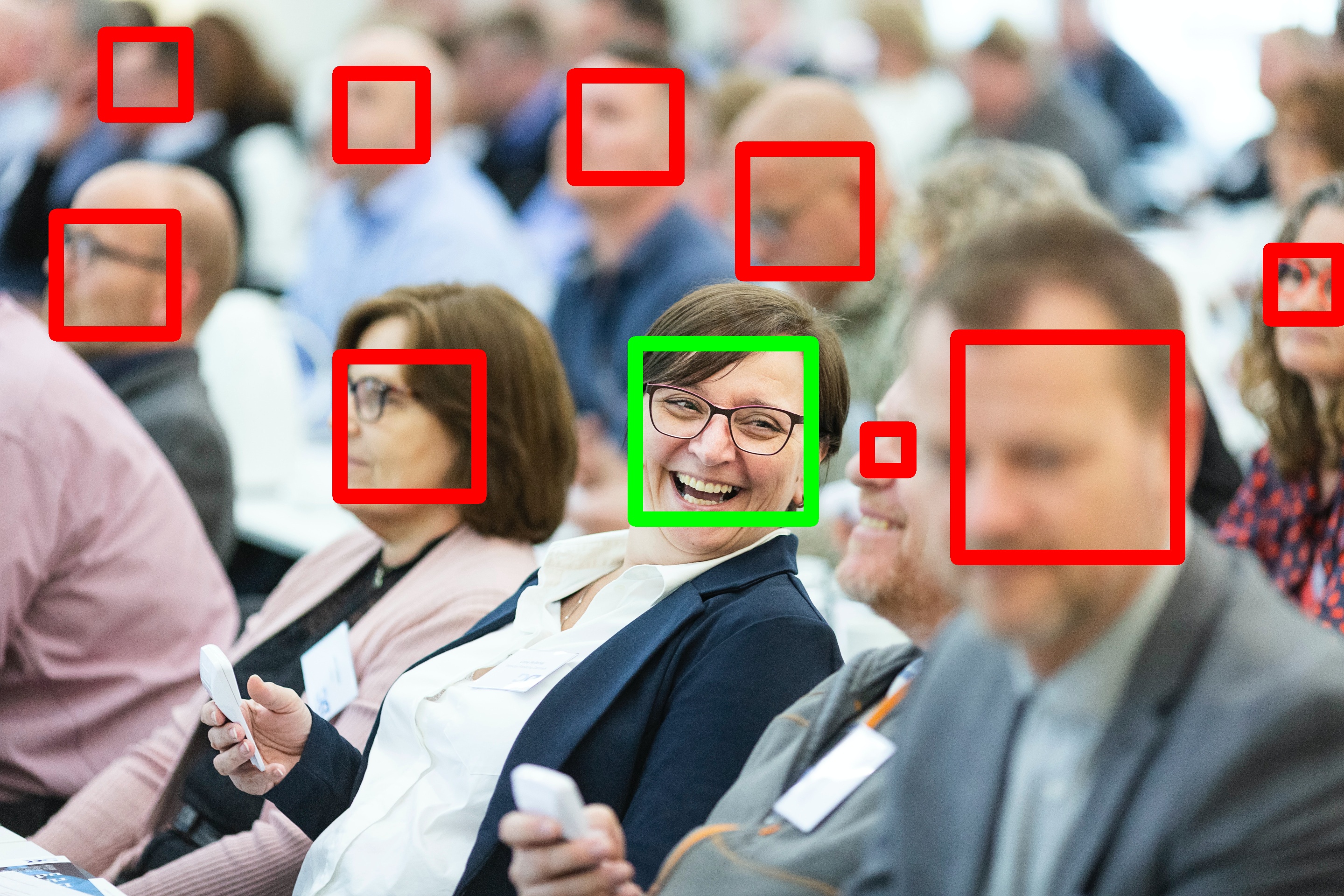}
    \caption{An example of automatic main character recognition (\cite{seker2021maincharacter}).(Image source: \textit{https://www.pexels.com})}
    \label{fig:maincharacter}
\end{figure}

Machine learning methods offer a way to carry out large-scale salience analysis (AICE \ref{var:saliency}) in an objective manner and to combine the salience analysis with other AICE variables or with the textual content to analyze how certain topics are illustrated. Recognizing and analysing the main character can be seen as a subtask of salience analysis and it is traditionally one of the basic tasks - and variables - in wide variety of photographic studies in the fields of social sciences and humanities. A recent work (\cite{seker2021maincharacter}) demonstrated that it is possible to automatically recognize the main character with a high accuracy in different kinds of photos as illustrated in Fig. \ref{fig:maincharacter}.

\subsection{Appearance-based search and analysis}
\label{ssec:appearancebased}

It is worth emphasizing that the new opportunities machine learning-enabled analysis provide extend beyond the existing predefined categories. When deep neural networks are trained, they internally learn to form generalized representations of the images that allow them to classify also previously unseen images. These representations can be also extracted and used for searching or grouping images based on their visual similarity (AICE \ref{var:similarity}) as also discussed by \textcite{arnold2019distant} with respect to \emph{distant viewing} of the databases.

Maybe the most straightforward way to exploit these internal representations is \emph{content-based image retrieval} (\cite{tzelepi2018cbir}) via \emph{query by example}. This means that instead of searching for a particular category, it is possible to search for images that are most similar to a particular query image or a group of query images (AICE \ref{var:query}). Here, it should be noted that the similarity always depends on the task the training was performed for. For example, if a network was trained for recognizing faces, it will learn to extract representations that help it in this particular task. These representations can be then used for \emph{query by face}, where the query image can depict any (unknown) face and the network can find the most similar faces. In the same manner, the representations learned by networks trained for the corresponding tasks can be used, for example, for querying by objects, environments, or poses. An illustrative example of \emph{query by pose} is provided in Fig. \ref{fig:query}, where the top image is used for querying and possible query results with similar poses are shown below. As the similarity of the photos is measured by the similarity of poses in this example, other image content can be very different, while the photo is still considered as a close match to the query image.
A more practical example of using such an approach in photographic studies was provided by \textcite{wevers2020visual} in their second task for querying the advertisements based on their abstract visual aspects.

\begin{figure}[t]
    \centering
    \includegraphics[width=1\columnwidth]{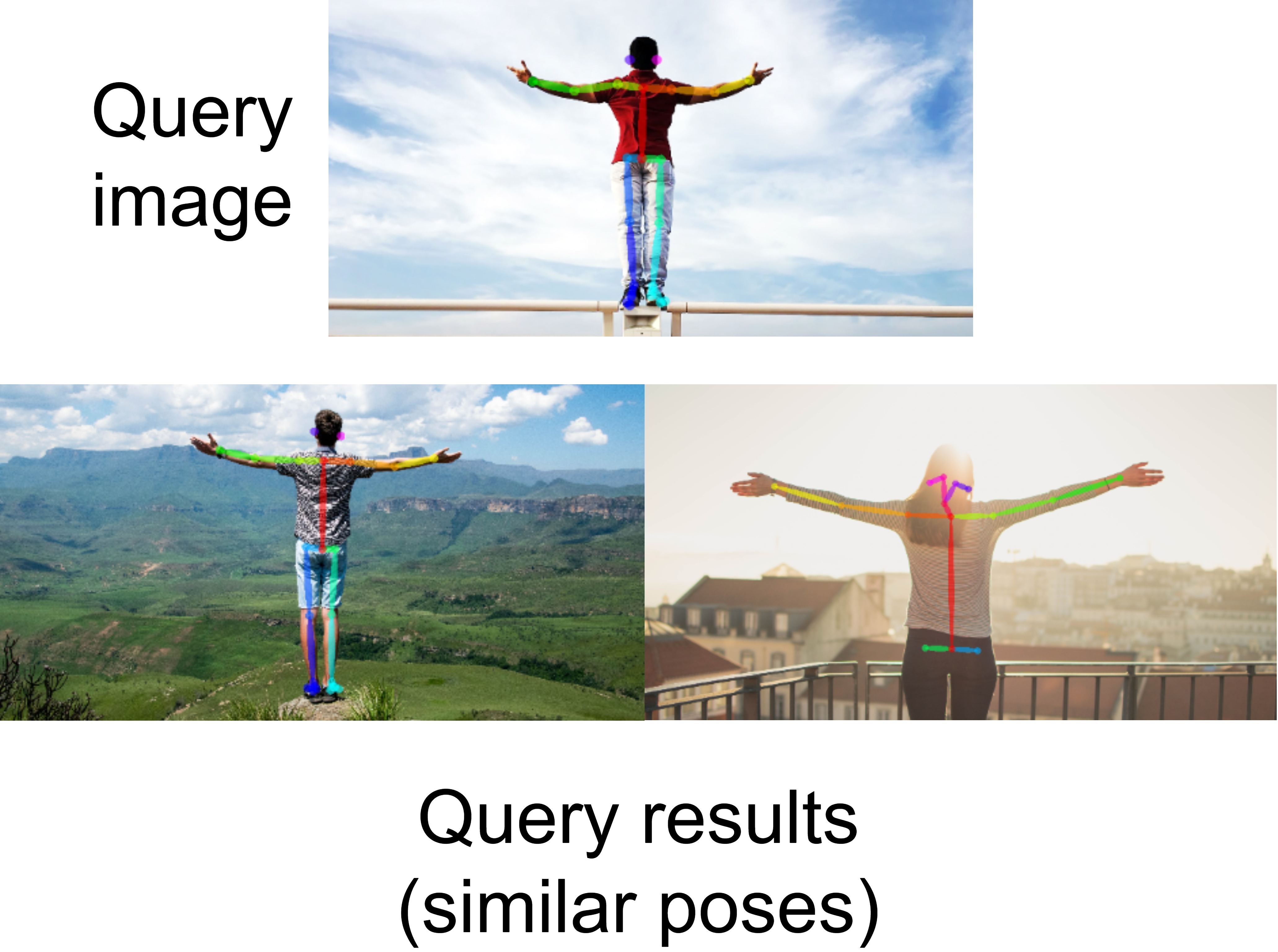}
    \caption{An illustration of content-based image retrieval in the case of query by pose.(Image source: \textit{https://www.pexels.com})}
    \label{fig:query}
\end{figure}

\begin{figure*}[t]
    \centering
    \includegraphics[width=0.9\textwidth]{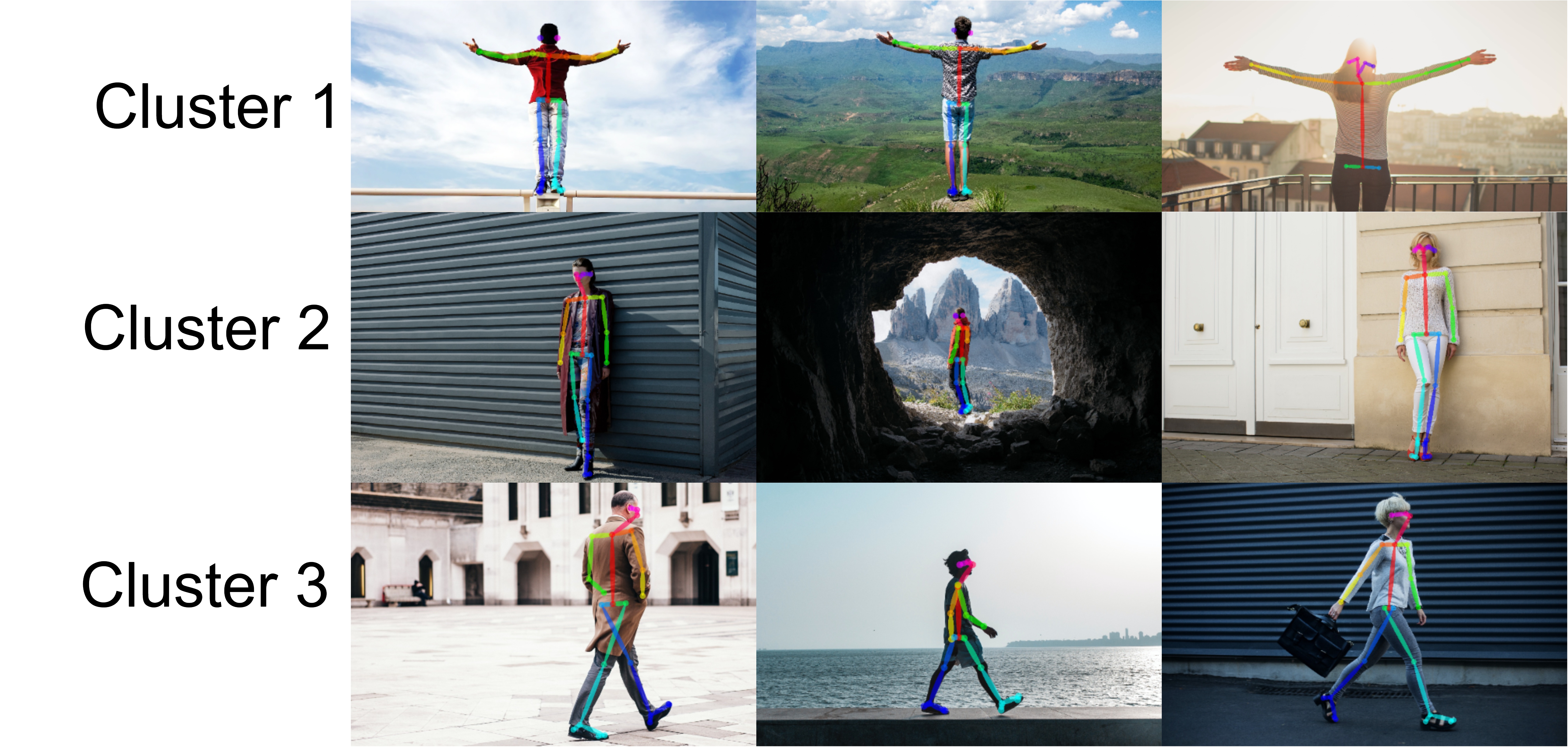}
    \caption{An illustration of clustering images by the similarity of poses.(Image source: \textit{https://www.pexels.com})}
    \label{fig:clustering}
\end{figure*}

Furthermore, the network representation can be used for grouping the images based on their similarity (AICE \ref{var:clustering}). This can allow revealing such similarities and groupings of data that the researcher would not be able to look for. There are also tools for visualizing such found similarities and groupings as demonstrated in a recent paper (\cite{chumachenko2020ww2}), where the similarities between different photographers were illustrated based on clustering their photos. A simplified illustration of clustering images into groups based on the similarity of poses is provided in Fig.~\ref{fig:clustering}, where the clustering algorithm have identified three distinct pose groups. It should be noted that in real photo collections the clusters are rarely so clear and distinct, but the images represent a more continuous distribution of poses (or any other feature used for clustering) making the clusters fuzzier by nature.  

As mentioned in Section~\ref{ssec:actions}, tools for \ref{var:query}, i.e. querying by example, could be used to find out whether an interesting gesture or pose appears repeatedly in a photo collection, and tools for \ref{var:clustering}, i.e. clustering based on visual similarity, could be used to divide action and poses automatically into groups containing similar gestures. These techniques are most suitable for studying such concepts where the text-based descriptions are cumbersome, but the visual similarity of the studied categories is high. The appearance-based techniques can also help avoid the difficulty of predefining the pose/action categories of interest and may also reveal similarities that could be very difficult to observe manually.
 
\section{Challenges and risks of the automatic approach}
\label{sec:risks}

In this paper, we have extensively discussed the new opportunities and directions machine learning-based methods open in photographic studies. However, it is also good to keep in mind that there are also some challenges and risks related to the new tools. In this section, we review some of the challenges, and also give our suggestions for mitigating them.

Maybe the most practical problem is that for many researchers in humanities having non-technical background, it may be difficult to find and apply suitable machine learning tools. Even though most of the tools discussed in this paper are publicly available, it requires some expertise to understand which methods are the most suitable ones, how to apply pretrained models, and how to adjust them for specific tasks. On the other hand, there are still interesting novel challenges also from the machine learning point of view. Therefore, we believe that interdisciplinary research between humanistic and machine learning researchers is the most fruitful solution to the practical issues and novel technical challenges. Furthermore, we believe that publishing the used algorithms in an easy-to-use format (e.g., as in \cite{seker2021maincharacter}) can help other researchers in their work and gradually lead towards easier application of the tools as well as easier comparison with earlier works.  

It is important to keep in mind in the analysis of the results that the machine learning models are not error free, and the harder the task, the lower the expected accuracy. Also different machine learning models can have very different performance in the same task, and it is important to keep in mind that a low performance may result from an unsuitable or poorly employed model, while other machine learning models might perform better. For the same reason, it is important to be aware of the employed models in any comparative studies and not directly compare the results of studies conducted using different models for the same task, because this might lead to biased or erroneous conclusions.

Some of the errors made by machine learning models may appear unexplainable or even absurd due to the \emph{black box nature} of the algorithms, which refers to the difficulty of explaining why a model gives a certain output, e.g., a classification decision (\cite{rudin2019stop}). The black box nature is an enormous problem in the fields such as medicine or autonomous driving, where a single critical error may lead to life-threatening situations. In the humanistic studies, such individual errors are much less severe, but the possibility of errors should be still considered in the analysis. For this reason, we see machine learning algorithms most suitable for searching suitable research material among very large photo collecting and for conducting large-scale comparative studies. As long as errors can be expected to occur at approximately the same rate on each dataset, the results can indicate true findings.  At the same time, explainable machine learning (\cite{tjoa2020survey}) is expected to develop rapidly due to the pressure in the fields with huge economic value, and these developments can then benefit also the future humanistic photographic studies.

A more challenging problem than individual errors are biased results or analysis. In their survey on biases and fairness in machine learning, \textcite{mehrabi2021survey} define fairness in the context of decision making as ``\emph{absence of any prejudice or favoritism toward an individual or group based on their inherent or acquired characteristics}'' and list several different sources of bias in machine learning applications. Most of the machine learning algorithms are neutral by nature, but they learn based on the training data shown to them and any bias in the data can be adopted or even emphasized by the model (\cite{wang2020revise}). For example, if a facial expression recognition algorithm is trained using mainly photos of young white people, it is natural that it will be less accurate with old and/or people of colour. Many of the data-related biases have a humanistic or societal nature, and more collaboration between humanistic and machine learning researchers will be needed to better understand how these biases are reflected in the machine learning-based analysis, not only in humanistic studies but in various applications across the society. Besides the training data, also the design of the machine learning models may introduce some biases. For example, in our recent social distance estimation algorithm (\cite{seker2021distance}), we assume that people’s torsos are upright and underestimates the distances if this assumption is violated. This happens more commonly, when people are sitting, which, in turn, most likely leads to more underestimation in indoor images. To tackle different biases, we further emphasize the importance of close collaboration between humanistic and machine learning researcher as well as open publication of the applied models and data whenever possible without privacy violations (\cite{ borgesius2015open}). This will allow critical evaluation of the results and biases in the future studies and gradually lead to better understanding of possible pitfalls. 

When machine learning becomes more widely-used in humanistic studies, it is expected that the availability of high-quality datasets will encourage research in the related themes. While the opportunities can open exciting new directions in humanistic studies, it may become necessary to consider, whether this will dictate the topics and research question setting too much leading to increased inequality, e.g., by favoring majority cultures instead of minority cultures as research topics. In a similar way, if a certain category of interest is missing from the available training data, it may be tempting to just omit the category from the research, leading to altered research directions. To avoid such problems, humanistic field should not see machine learning as a ready-to-use tool, but also take responsibility in annotating large-scale image collections with the concepts relevant for the subsequent humanistic studies. To this end, the development of both faster and easier annotation tools and methods for incremental learning that allow adding some categories to already trained algorithms (e.g., \textcite{roy2020tree}) are important for future photographic studies.

\section{Conclusions}
\label{sec:conclusion}

Introduction of novel machine learning methods has potential to both significantly reduce the time needed for selecting the suitable material and at the same time to give a more detailed picture of the quantitative features of very large image collections. The data-driven analysis and ability to automatically search and annotate content of images can remove the most laborious tasks in visual content analysis. This may also lead to decreasing the amount of errors in the observation or note taking process. These changes will help to shift the workflow from simple manual tasks to more demanding stages. This includes analyzing and examining the databases, performing a qualitative phase where the researcher seeks to answer ‘what the data mean’ (\cite[22]{bell2004content}). It will also enable the use of much larger sets of photos than has been traditionally possible. As each photo can be annotated with these new tools with detailed information of its content and structural patterns and locations, it will enable creating much more sophisticated research settings. Overall, the novel machine learning techniques will allow to renew at least the quantitative part of photographic studies in the near future.

In this paper, we proposed a holistic framework for machine learning-based image analysis called Automatic Image Content Extraction (AICE). To make the traditional visual content analysis methodologies compatible with the novel machine learning techniques, we reformulated and expanded the traditional framework by adding several variables and suggesting suitable values for each variable. We also considered the difficulty of automatic extraction of each variable with the current state-of-the-art machine learning models and linked the variables to existing machine learning techniques.

The proposed framework can be applied in several domains in humanities and social sciences. We provided multiple examples on how machine learning techniques applied according to the proposed AICE framework could have enhanced earlier photographic studies and illustrated the novel opportunities they open for future studies. We also discussed the main challenges in adopting the automatic approach and suggested solutions to these challenges. As the main solution, we encourage expanding collaboration between humanistic and machine learning researchers. In addition, we believe that humanistic field should not see machine learning as a ready-to-use tool, but also take responsibility in annotating large-scale image collections with the concepts relevant for the subsequent humanistic studies.

While this paper and our proposed AICE framework focus solely on images, it is worth noting that the novel opportunities opened by the advancement of machine learning techniques are wider. In particular, different \emph{multi-modal} techniques that simultaneously analyze different sources of information, such as news texts, graphical elements, or page layout along with the images or videos consisting of both visual and audio data, can open additional interesting opportunities for the future humanistic and social science studies.

\section*{Acknowledgment}

We would like to acknowledge support from Intelligent Society Platform funded by Academy of Finland, project “Improving Public Accessibility of Large Image Archives”.

\printbibliography

@inproceedings{altinigne2020height,
  title={Height and Weight Estimation from Unconstrained Images},
  author={Altinigne, Can Yilmaz and Thanou, Dorina and Achanta, Radhakrishna},
  booktitle={IEEE International Conference on Acoustics, Speech and Signal Processing (ICASSP)},
  pages={2298--2302},
  year={2020},
}

@article{araujo2020automated,
  title={Automated visual content analysis (AVCA) in communication research: A protocol for large scale image classification with pre-trained computer vision models},
  author={Araujo, Theo and Lock, Irina and van de Velde, Bob},
  journal={Communication Methods and Measures},
  volume={14},
  number={4},
  pages={239--265},
  year={2020},
  publisher={Taylor \& Francis}
}

@article{arnold2019distant,
  title={Distant viewing: analyzing large visual corpora},
  author={Arnold, Taylor and Tilton, Lauren},
  journal={Digital Scholarship in the Humanities},
  volume={34, Supplement 1},
  pages={i3--i16},
  year={2019},
  publisher={Oxford University Press}
}

@inproceedings{arnold2020enriching,
  title={Enriching historic photography with structured data using image region segmentation},
  author={Arnold, Taylor and Tilton, Lauren},
  booktitle={Proceedings of the 1st International Workshop on Artificial Intelligence for Historical Image Enrichment and Access},
  pages={1--10},
  year={2020}
}

@article{bell2002goffman,
  title={Goffman’s gender advertisements revisited: Combining content analysis with semiotic analysis},
  author={Bell, Philip and Milic, Marko},
  journal={Visual communication},
  volume={1},
  number={2},
  pages={203--222},
  year={2002},
  publisher={Sage Publications Sage CA: Thousand Oaks, CA}
}

@article{bell2004content,
  title={Content analysis of visual images},
  author={Bell, Philip},
  journal={Handbook of visual analysis},
  year={2004},
  publisher={Thousand Oaks, CA}
}

@article{berelson1952content,
  title={Content analysis in communication research.},
  author={Berelson, Bernard},
  year={1952},
  publisher={Free press}
}

@article{bodker2018journalism,
  title={Journalism history and digital archives},
  author={B{\o}dker, Henrik},
  journal={Digital Journalism},
  volume={6},
  number={9},
  pages={1113--1120},
  year={2018},
  publisher={Taylor \& Francis}
}

@article{borgesius2015open,
  title={Open data, privacy, and fair information principles: Towards a balancing framework},
  author={Borgesius, Frederik Zuiderveen and Gray, Jonathan and van Eechoud, Mireille},
  journal={Berkeley Technology Law Journal},
  volume={30},
  number={3},
  pages={2073--2131},
  year={2015},
  publisher={JSTOR}
}

@article{braun2019eurocity,
  title={Eurocity persons: A novel benchmark for person detection in traffic scenes},
  author={Braun, Markus and Krebs, Sebastian and Flohr, Fabian and Gavrila, Dariu M},
  journal={IEEE Transactions on Pattern Analysis and Machine Intelligence},
  volume={41},
  number={8},
  pages={1844--1861},
  year={2019},
  publisher={IEEE}
}

@article{broersma2018exploring,
  title={Exploring Machine Learning to Study the Long-Term Transformation of News: Digital newspaper archives, journalism history, and algorithmic transparency},
  author={Broersma, Marcel and Harbers, Frank},
  journal={Digital Journalism},
  volume={6},
  number={9},
  pages={1150--1164},
  year={2018},
  publisher={Taylor \& Francis}
}

@article{cao2019openpose,
  title={{OpenPose: realtime multi-person 2D pose estimation using Part Affinity Fields}},
  author={Cao, Zhe and Hidalgo, Gines and Simon, Tomas and Wei, Shih-En and Sheikh, Yaser},
  journal={IEEE Transactions on Pattern Analysis and Machine Intelligence},
  volume={43},
  number={1},
  pages={172--186},
  year={2019}
}

@article{cerku2019applied,
  title={Applied visual anthropology in the progressive era: The influence of Lewis Hine’s child labor photographs},
  author={Cerku, Ashley},
  journal={Visual Anthropology},
  volume={32},
  number={3-4},
  pages={221--239},
  year={2019},
  publisher={Taylor \& Francis}
}

@article{collier2004approaches,
  title={Approaches to analysis in visual anthropology},
  author={Collier, Malcolm},
  journal={Handbook of visual analysis},
  pages={35--60},
  year={2004},
  publisher={London}
}

@article{chumachenko2020ww2,
  title={{Machine learning based analysis of Finnish World War II photographers}},
  author={Chumachenko, Kateryna and M{\"a}nnist{\"o}, Anssi and Iosifidis, Alexandros and Raitoharju, Jenni},
  journal={IEEE Access},
  volume={8},
  pages={144184--144196},
  year={2020}
}

@article{cristani2020distance,
  title={The Visual Social Distancing Problem},
  author={Cristani, M. and Bue, A. D. and Murino, V. and Setti, F. and Vinciarelli, A.},
  journal={IEEE Access},
  volume={8},
  pages={126876-126886},
  year={2020}
}

@article{dong2021action,
  title={Knowledge memorization and generation for action recognition in still images},
  author={Dong, Jian and Yang, Wankou and Yao, Yazhou and Porikli, Fatih},
  journal={Pattern Recognition},
  volume={120},
  pages={108188},
  year={2021},
}

@book{emmison2000researching,
  title={Researching the Visual: Images, Objects, Contexts and Interactions in Social and Cultural Inquiry},
  author={Emmison, Michael and Smith, Philip},
  year={2000},
  publisher = {London: SAGE},
}

@article{fiorucci2020machine,
  title={Machine learning for cultural heritage: A survey},
  author={Fiorucci, Marco and Khoroshiltseva, Marina and Pontil, Massimiliano and Traviglia, Arianna and Del Bue, Alessio and James, Stuart},
  journal={Pattern Recognition Letters},
  volume={133},
  pages={102--108},
  year={2020},
}

@inproceedings{fu2019segmentation,
  title={Dual attention network for scene segmentation},
  author={Fu, Jun and Liu, Jing and Tian, Haijie and Li, Yong and Bao, Yongjun and Fang, Zhiwei and Lu, Hanqing},
  booktitle={IEEE/CVF Conference on Computer Vision and Pattern Recognition (CVPR)},
  pages={3146--3154},
  year={2019}
}

@book{goffman1979gender,
  title={Gender advertisements},
  author={Goffman, Erving},
  year={1979},
  publisher={London: Macmillan Press}
}

@book{gonzalez2018digital,
  title={Digital image processing},
  author={Gonzalez, Rafael C and Woods, Richard E},
  year={2018},
  edition = {4},
  publisher={Pearson}
}

@article{grady2007advertising,
  title={Advertising images as social indicators: depictions of blacks in LIFE magazine, 1936--2000},
  author={Grady, John},
  journal={Visual studies},
  volume={22},
  number={3},
  pages={211--239},
  year={2007},
  publisher={Taylor \& Francis}
}

@article{griffin1995picturing,
  title={Picturing the Gulf War: constructing an image of war in Time, Newsweek, and US News \& World Report},
  author={Griffin, Michael and Lee, Jongsoo},
  journal={Journalism \& Mass Communication Quarterly},
  volume={72},
  number={4},
  pages={813--825},
  year={1995},
  publisher={SAGE Publications Sage CA: Los Angeles, CA}
}

@article{griffin2004picturing,
  title={{Picturing America’s ‘War on Terrorism’in Afghanistan and Iraq: Photographic motifs as news frames}},
  author={Griffin, Michael},
  journal={Journalism},
  volume={5},
  number={4},
  pages={381--402},
  year={2004},
  publisher={Sage Publications Sage CA: Thousand Oaks, CA}
}

@book{hall1966hidden,
  title={The hidden dimension},
  author={Hall, Edward T.},
  year={1966},
  publisher={Garden City, NY: Doubleday}
}

@inproceedings{ion2019nudity,
  title={Application of Image Classification for Fine-Grained Nudity Detection},
  author={Ion, Cristian and Minea, Cristian},
  booktitle={International Symposium on Visual Computing},
  pages={3--15},
  year={2019},
}

@article{jay2002cultural,
  title={Cultural relativism and the visual turn},
  author={Jay, Martin},
  journal={Journal of visual culture},
  volume={1},
  number={3},
  pages={267--278},
  year={2002},
  publisher={Sage Publications Sage CA: Thousand Oaks, CA}
}

@InProceedings{kellnhofer2019gaze,
author = {Kellnhofer, Petr and Recasens, Adria and Stent, Simon and Matusik, Wojciech and Torralba, Antonio},
title = {Gaze360: Physically Unconstrained Gaze Estimation in the Wild},
booktitle = {IEEE/CVF International Conference on Computer Vision (ICCV)},
year = {2019}
}

@inproceedings{kendall2017camerapose,
  title={Geometric loss functions for camera pose regression with deep learning},
  author={Kendall, Alex and Cipolla, Roberto},
  booktitle={IEEE conference on Computer Vision and Pattern Recognition (CVPR)},
  pages={5974--5983},
  year={2017}
}

@inbook{kozol2009life, 
 title={Life Magazine Photographs}, 
 booktitle={Visual Communication Research Designs},
 publisher={NY: Routledge}, 
 author={Kozol, Wendy}, 
 year={2009}}

@inbook{kress1996reading, 
 place={Geelong, Vic.}, 
 title={Reading images: The grammar of visual design. 1st edition.}, 
 publisher={New York: Roudledge}, 
 author={Kress, Gunther R. and van Leeuwen, Theo}, 
 year={1996}}

@inbook{kress2021reading, 
 title={Reading images: The grammar of visual design. 4th edition.}, 
 publisher={London: Roudledge}, 
 author={Kress, Gunther R. and van Leeuwen, Theo}, 
 year={2021}}

@article{krizhevsky2012imagenet,
  title={Imagenet classification with deep convolutional neural networks},
  author={Krizhevsky, Alex and Sutskever, Ilya and Hinton, Geoffrey E},
  journal={Advances in Neural Information Processing Systems (NIPS)},
  volume={25},
  pages={1097--1105},
  year={2012}
}

@incollection{kuhn2018metadata,
  title={Images on the Move: Analytics for a Mixed Methods Approach},
  author={Kuhn, Virginia},
  booktitle={The Routledge Companion to Media Studies and Digital Humanities},
  pages={300--309},
  year={2018},
  publisher={Routledge}
}

@article{lecun2015deep,
  title={Deep learning},
  author={LeCun, Yann and Bengio, Yoshua and Hinton, Geoffrey},
  journal={Nature},
  volume={521},
  number={7553},
  pages={436--444},
  year={2015},
  publisher={Nature Publishing Group}
}

@InProceedings{li2015saliency,
author = {Li, Guanbin and Yu, Yizhou},
title = {Visual Saliency Based on Multiscale Deep Features},
booktitle = {IEEE Conference on Computer Vision and Pattern Recognition (CVPR)},
year = {2015}
}

@article{li2020expression,
  title={Deep facial expression recognition: A survey},
  author={Li, Shan and Deng, Weihong},
  journal={IEEE Transactions on Affective Computing},
  year={2020},
}

@inproceedings{li2019hoi,
  title={Transferable interactiveness knowledge for human-object interaction detection},
  author={Li, Yong-Lu and Zhou, Siyuan and Huang, Xijie and Xu, Liang and Ma, Ze and Fang, Hao-Shu and Wang, Yanfeng and Lu, Cewu},
  booktitle={IEEE/CVF Conference on Computer Vision and Pattern Recognition (CVPR)},
  pages={3585--3594},
  year={2019}
}

@inproceedings{liu2016fashion,
  title={Deepfashion: Powering robust clothes recognition and retrieval with rich annotations},
  author={Liu, Ziwei and Luo, Ping and Qiu, Shi and Wang, Xiaogang and Tang, Xiaoou},
  booktitle={IEEE Conference on Computer Vision and Pattern Recognition (CVPR)},
  pages={1096--1104},
  year={2016}
}

@article{liu2020detection,
  title={Deep learning for generic object detection: A survey},
  author={Liu, Li and Ouyang, Wanli and Wang, Xiaogang and Fieguth, Paul and Chen, Jie and Liu, Xinwang and Pietik{\"a}inen, Matti},
  journal={International Journal of Computer Vision},
  volume={128},
  number={2},
  pages={261--318},
  year={2020},
}

@article{manovich2013visualizing,
  title={Visualizing change: Computer graphics as a research method},
  author={Manovich, Lev and Douglass, Jeremy},
  journal={Imagery in the 21st Century},
  pages={315--38},
  year={2013},
  publisher={MIT Press London}
}

@incollection{manovich2012compare,
  title={How to compare one million images?},
  author={Manovich, Lev},
  booktitle={Understanding digital humanities},
  pages={249--278},
  year={2012},
}

@article{mannisto2004luottamus,
  title={Luottamus katosi savuna ilmaan. Journalismikritiikin vuosikirja 2004.},
  author={Anssi M{\"a}nnist{\"o}},
  journal={Media \& Viestintä},
  volume={27},
  number = {1},
  year={2004},
}

@journal{matzen2017streetstyle,
      title={StreetStyle: Exploring world-wide clothing styles from millions of photos}, 
      author={Kevin Matzen and Kavita Bala and Noah Snavely},
      year={2017},
      journal ={arXiv 1706.01869},
}

@article{mehrabi2021survey,
  title={A survey on bias and fairness in machine learning},
  author={Mehrabi, Ninareh and Morstatter, Fred and Saxena, Nripsuta and Lerman, Kristina and Galstyan, Aram},
  journal={ACM Computing Surveys (CSUR)},
  volume={54},
  number={6},
  pages={1--35},
  year={2021},
  publisher={ACM New York, NY, USA}
}

@article{memon2020handwritten,
  title={Handwritten optical character recognition (OCR): A comprehensive systematic literature review (SLR)},
  author={Memon, Jamshed and Sami, Maira and Khan, Rizwan Ahmed and Uddin, Mueen},
  journal={IEEE Access},
  volume={8},
  pages={142642--142668},
  year={2020},
}

@article{min2018clustering,
  title={A survey of clustering with deep learning: From the perspective of network architecture},
  author={Min, Erxue and Guo, Xifeng and Liu, Qiang and Zhang, Gen and Cui, Jianjing and Long, Jun},
  journal={IEEE Access},
  volume={6},
  pages={39501--39514},
  year={2018},
}

@article{morse2014covering,
  title={Covering the Dead: Death images in Israeli newspapers—ethics and praxis},
  author={Morse, Tal},
  journal={Journalism Studies},
  volume={15},
  number={1},
  pages={98--113},
  year={2014},
  publisher={Taylor \& Francis}
}

@article{nicholson2013digital,
  title={The Digital Turn: Exploring the methodological possibilities of digital newspaper archives},
  author={Nicholson, Bob},
  journal={Media History},
  volume={19},
  number={1},
  pages={59--73},
  year={2013},
}

@article{parry2020quantitative,
  title={Quantitative content analysis of the visual},
  author={Parry, Katy},
  journal={The SAGE Handbook of Visual Research Methods},
  pages={352--366},
  year={2020}
}

@article{pauwels2000taking,
  title={Taking the visual turn in research and scholarly communication key issues in developing a more visually literate (social) science},
  author={Pauwels, Luc},
  journal={Visual studies},
  volume={15},
  number={1},
  pages={7--14},
  year={2000},
  publisher={Taylor \& Francis}
}

@article{pauwels2020integrated,
  title={An integrated conceptual framework for visual social},
  author={Pauwels, Luc},
  journal={The SAGE Handbook of Visual Research Methods},
  year={2020},
}

@book{pauwels2020sage,
  title={The SAGE handbook of visual research methods},
  author={Pauwels, Luc and Mannay, Dawn},
  year={2020},
  publisher={Sage}
}

@misc{1993reading,
  title = {Reading National Geographic},
  note = {Book brochure}, 
  year = {1993},
  url = {https://press.uchicago.edu/ucp/books/book/chicago/R/bo3697068.html},
  urldate = {2021-12-09} 
  }

@book{riffe2019analyzing,
  title={Analyzing media messages: Using quantitative content analysis in research},
  author={Riffe, Daniel and Lacy, Stephen and Watson, Brendan R and Fico, Frederick},
  year={2019},
  publisher={NY: Routledge}
}

@article{rodriguez2017age,
  title={Age and gender recognition in the wild with deep attention},
  author={Rodriguez, Pau and Cucurull, Guillem and Gonfaus, Josep M and Roca, F Xavier and Gonzalez, Jordi},
  journal={Pattern Recognition},
  volume={72},
  pages={563--571},
  year={2017},
}

@book{rose2016visual,
  title={Visual methodologies: An introduction to researching with visual materials},
  author={Rose, Gillian},
  year={2016},
  edition = {4},
  publisher={London: SAGE}
}

@article{roy2020tree,
  title={Tree-CNN: a hierarchical deep convolutional neural network for incremental learning},
  author={Roy, Deboleena and Panda, Priyadarshini and Roy, Kaushik},
  journal={Neural Networks},
  volume={121},
  pages={148--160},
  year={2020},
  publisher={Elsevier}
}

@article{rudin2019stop,
  title={Stop explaining black box machine learning models for high stakes decisions and use interpretable models instead},
  author={Rudin, Cynthia},
  journal={Nature Machine Intelligence},
  volume={1},
  number={5},
  pages={206--215},
  year={2019},
  publisher={Nature Publishing Group}
}

@inproceeding{seker2021maincharacter,
      title={Automatic Main Character Recognition for Photographic Studies}, 
      author={Mert Seker and Anssi Männistö and Alexandros Iosifidis and Jenni Raitoharju},
      year={2021},
      booktitle={IEEE International Workshop on Multimedia Signal Processing (MMSP)},
}

@journal{seker2021distance,
      title={Automatic Social Distance Estimation From Images: Performance Evaluation, Test Benchmark, and Algorithm}, 
      author={Mert Seker and Anssi Männistö and Alexandros Iosifidis and Jenni Raitoharju},
      year={2021},
      journal={arXiv 2103.06759},
}

@article{sherren2017digital,
  title={Digital archives, big data and image-based culturomics for social impact assessment: opportunities and challenges},
  author={Sherren, Kate and Parkins, John R and Smit, Michael and Holmlund, Mona and Chen, Yan},
  journal={Environmental Impact Assessment Review},
  volume={67},
  pages={23--30},
  year={2017},
}

@article{stergiou2019interaction,
  title={Analyzing human--human interactions: A survey},
  author={Stergiou, Alexandros and Poppe, Ronald},
  journal={Computer Vision and Image Understanding},
  volume={188},
  pages={102799},
  year={2019},
}

@article{temelova2011daily,
  title={Daily street life in the inner city of Prague under transformation: the visual experience of socio-spatial differentiation and temporal rhythms},
  author={Temelov{\'a}, Jana and Nov{\'a}k, Jakub},
  journal={Visual Studies},
  volume={26},
  number={3},
  pages={213--228},
  year={2011},
  publisher={Taylor \& Francis}
}

@article{tjoa2020survey,
  title={A survey on explainable artificial intelligence (xai): Toward medical xai},
  author={Tjoa, Erico and Guan, Cuntai},
  journal={IEEE Transactions on Neural Networks and Learning Systems},
  year={2020},
  publisher={IEEE}
}

@article{tzelepi2018cbir,
  title={Deep convolutional learning for content based image retrieval},
  author={Tzelepi, Maria and Tefas, Anastasios},
  journal={Neurocomputing},
  volume={275},
  pages={2467--2478},
  year={2018},
  publisher={Elsevier}
}

@article{vaisanen2021exploring,
  title={Exploring human--nature interactions in national parks with social media photographs and computer vision},
  author={V{\"a}is{\"a}nen, Tuomas and Heikinheimo, Vuokko and Hiippala, Tuomo and Toivonen, Tuuli},
  journal={Conservation Biology},
  volume={35},
  number={2},
  pages={424--436},
  year={2021},
  publisher={Wiley Online Library}
}

@article{wang2018events,
  title={Transferring deep object and scene representations for event recognition in still images},
  author={Wang, Limin and Wang, Zhe and Qiao, Yu and Van Gool, Luc},
  journal={International Journal of Computer Vision},
  volume={126},
  number={2},
  pages={390--409},
  year={2018},
}

@article{wang2021face,
  title={Deep face recognition: A survey},
  author={Wang, Mei and Deng, Weihong},
  journal={Neurocomputing},
  volume={429},
  pages={215--244},
  year={2021},
  publisher={Elsevier}
}

@inproceedings{wang2020revise,
  title={REVISE: A tool for measuring and mitigating bias in visual datasets},
  author={Wang, Angelina and Narayanan, Arvind and Russakovsky, Olga},
  booktitle={European Conference on Computer Vision},
  pages={733--751},
  year={2020},
  organization={Springer}
}

@article{wevers2020visual,
  title={The visual digital turn: Using neural networks to study historical images},
  author={Wevers, Melvin and Smits, Thomas},
  journal={Digital Scholarship in the Humanities},
  volume={35},
  number={1},
  pages={194--207},
  year={2020},

}

@article{wilmott2017politics,
  title={The politics of photography: Visual depictions of Syrian refugees in UK online media},
  author={Wilmott, Annabelle Cathryn},
  journal={Visual Communication Quarterly},
  volume={24},
  number={2},
  pages={67--82},
  year={2017},
}

@inproceedings{yu2017relationship,
  title={Visual relationship detection with internal and external linguistic knowledge distillation},
  author={Yu, Ruichi and Li, Ang and Morariu, Vlad I and Davis, Larry S},
  booktitle={IEEE International Conference on Computer Vision (CVPR)},
  pages={1974--1982},
  year={2017}
}

@article{zhao2018weather,
  title={A CNN--RNN architecture for multi-label weather recognition},
  author={Zhao, Bin and Li, Xuelong and Lu, Xiaoqiang and Wang, Zhigang},
  journal={Neurocomputing},
  volume={322},
  pages={47--57},
  year={2018},
}

@article{zhou2017places,
  title={Places: A 10 million image database for scene recognition},
  author={Zhou, Bolei and Lapedriza, Agata and Khosla, Aditya and Oliva, Aude and Torralba, Antonio},
  journal={IEEE Transactions on Pattern Analysis and Machine Intelligence},
  volume={40},
  number={6},
  pages={1452--1464},
  year={2017},
}

\end{document}